\documentclass[11pt]{article}

\usepackage[utf8]{inputenc}
\usepackage[T1]{fontenc}
\usepackage{lmodern}
\usepackage[margin=1in]{geometry}
\usepackage{amsmath,amssymb,amsthm}
\usepackage{mathtools}
\usepackage{booktabs}
\usepackage{array}
\usepackage{tabularx}
\usepackage{ragged2e}
\newcolumntype{L}{>{\RaggedRight\arraybackslash}X}
\usepackage{enumitem}
\usepackage{microtype}
\usepackage{tikz}
\usetikzlibrary{arrows.meta, positioning, fit, backgrounds}
\usepackage[numbers,sort&compress]{natbib}
\usepackage[hidelinks]{hyperref}
\usepackage{cleveref}

\theoremstyle{definition}
\newtheorem{definition}{Definition}
\newtheorem{designgoal}{Design Goal}

\theoremstyle{plain}
\newtheorem{claim}{Claim}
\newtheorem{invariant}{Invariant}
\newtheorem{principle}{Principle}
\newtheorem{adjalg}{Algorithm}

\title{\textbf{A Five-Plane Reference Architecture for\\ Runtime Governance of Production AI Agents}}
\author{Krti Tallam\\\texttt{krti@kamiwaza.ai}}
\date{}

\begin{document}
\maketitle
\begin{abstract}
Enterprise security was built to govern data boundaries: the protected surface was data at rest and in transit, and the controls (access control, data-loss prevention, perimeter inspection) governed crossings of that boundary. Production AI agents dissolve this assumption. An agent reads context, calls tools, invokes connectors, and modifies systems of record on an enterprise's behalf, so risk moves inside the workflow, into sequences of individually-permitted actions that may transform a business process no one authorized. Existing policy engines do not extend to this regime: they evaluate request-time decisions against atomic principals, where agentic systems require stateful evaluation against composite principals whose authority attenuates through delegation chains. We present a reference architecture for the runtime governance of production agents, built from four composable primitives: a five-plane decomposition (a reasoning plane that adjudicates intent, plus four enforcement planes for network, identity, endpoint, and data that realize the decision); stop-anywhere mediation; composite principals with capability attenuation; and audit as a structured evidence substrate. We define a taxonomy of six interruption primitives that generalize allow and deny, state and argue for four correctness invariants, demonstrate the foreclosure of seven production-agent threats across five concrete workflows, and propose an evaluation framework that treats safety and utility as joint objectives. A reference implementation of the policy-engine core supplies measured evidence: attenuation correctness and evidence reconstructability hold on every trial, adjudication runs in single-digit microseconds, and the audit substrate's tamper-evidence behaves exactly as designed. We are explicit about scope: the architecture governs delegated action, not model behavior; its invariants are argued structurally, not formally proved; and the reference implementation validates the architecture's internal claims, leaving a full-system evaluation against a live agent benchmark as the invited next step.
\end{abstract}
\section{Introduction}

\subsection{The security definition is changing}

Enterprise security has, for two decades, organized itself around boundaries. The network perimeter defined a boundary between trusted and untrusted networks. The application boundary defined the points at which requests entered a system and were authorized. Data-loss prevention defined the boundary across which sensitive data must not pass. The controls that enforce these boundaries --- firewalls, access control, content inspection --- are mature, well-understood, and effective for the threat they were designed to address: the unauthorized crossing of a boundary by data or by a request.

Production AI agents change what must be protected. An agent deployed against an enterprise's systems of record does not merely cross boundaries; it acts. It reads context across connectors, forms plans, selects and invokes tools, and modifies records. The risk it introduces is not principally that data crosses a boundary --- though that risk remains --- but that the agent's sequence of decisions and actions transforms the enterprise's processes in ways the enterprise did not authorize. The protected surface is no longer infrastructure, data, or users in isolation. It is the interaction between human intent, the agent's generated decisions, the enterprise's context, and the agent's action against systems of record.

This shift has a direct consequence for the security stack. The controls built for boundary-crossing do not govern action sequences. Data-loss prevention inspects content in transit; it cannot evaluate whether the plan that produced the content was authorized. Access control authorizes a request against a principal; it cannot evaluate a delegation chain through which an agent acquired authority, or a sequence of individually-permitted actions whose composition is an exfiltration. The security definition is changing because the thing that must be governed has changed: from data crossing a boundary to action taken under delegated authority.

\subsection{Why existing policy engines do not extend}

The natural response to a new governance need is to extend the existing authorization stack. We argue that the existing stack does not extend, for five structural reasons.

First, existing engines are \textit{request-gated, not plan-aware}. Authorization systems --- role-based, attribute-based, or policy-as-code evaluators \cite{OPA,Cedar} --- evaluate one request at a time: may principal $p$ perform action $a$ on resource $r$? Agentic systems generate plans, and the policy-relevant question is not whether a single action is permitted but whether a \textit{sequence} of intended actions is permitted before any of them fires. Existing engines have no abstraction for intent.

Second, existing engines assume \textit{atomic principals}. They authorize against a single subject. Agentic systems have composite principals: a human delegates to a planner agent, which delegates to an executor agent, which invokes a tool that calls a downstream agent. Existing engines flatten this chain to one of its members --- typically the most recent caller --- discarding exactly the information needed to reason about whether the chain confers the authority being exercised.

Third, existing engines express \textit{absolute, not attenuated, authority}. They state rules in absolute terms; the foundational principle that delegated authority should be a strict subset of the delegator's --- attenuation, in the sense of object-capability theory \cite{Miller06,Birgisson14} --- is not native to them. They can deny, but they cannot natively express ``the executor agent holds the planner's capabilities, minus these.''

Fourth, existing engines are \textit{stateless}. By design \cite{OPAdesign}, a policy-as-code evaluator decides each request independently. Agentic governance requires decisions that depend on session state: an agent that has read confidential content earlier in a session may be forbidden from an external transmission later in the same session, even though the transmission is permitted in isolation. Encoding session state into a stateless evaluator's input is a correctness and performance pathology, not a solution.

Fifth, existing engines produce \textit{Boolean output}. Allow or deny is the output type. Production-agent governance requires richer outcomes --- modify the arguments, narrow the capability set, escalate to a human, defer pending a condition, roll back a committed effect --- that a Boolean evaluator cannot express.

These five gaps are not incidental; they are structural properties of engines designed for a different problem. The conclusion is not that the existing stack is wrong, but that it is insufficient: it governs request-time access against atomic principals, and production-agent governance requires plan-aware, stateful, attenuated, richly-output adjudication against composite principals.

\subsection{The reference architecture}

We present a reference architecture for the runtime governance of production agents. The architecture composes four primitives.

The first is a \textit{five-plane structural decomposition}. A production agent's action touches five planes: a reasoning plane, where intent is adjudicated against the composite principal and session state; and four infrastructure planes --- network, identity, endpoint, and data --- where the decision is realized using the enterprise's existing enforcement primitives. The reasoning plane decides once; the four planes realize the decision in coordination; a single composed evidence record binds the decision to its realizations. This decomposition replaces the prevailing federated model, in which each plane authorizes independently against partial context, with a composition in which adjudication happens once against full context.

The second is \textit{stop-anywhere mediation}. The reasoning plane is present at every meaningful point in the agent's execution pipeline --- seven such points, from plan formation to audit emission --- and its output vocabulary is a taxonomy of six interruption primitives (pause, escalate, narrow, modify, defer, rollback) that generalizes allow and deny. The system can intervene anywhere in the agent's loop, and its intervention is not limited to blocking.

The third is the \textit{composite principal with capability attenuation}. The principal against which the reasoning plane adjudicates is not an atomic identity but an explicit delegation chain, each step of which strictly attenuates the authority of the previous. The effective authority of any agent is the intersection of the capability sets along its chain, restricted to unexpired capabilities, and cryptographically bound to the chain itself. This makes the authority of any agent --- including a compromised one --- bounded by construction.

The fourth is \textit{audit as a structured evidence substrate}. Every decision produces a structured, complete, tamper-evident evidence record, reconstructible even under partial information. Audit is not logging tagged for compliance; it is evidence production designed for the auditor, the regulator, and the incident responder.

\subsection{Contributions}

This paper makes the following contributions.

\begin{enumerate}
\item \textbf{A five-plane reference architecture} for the runtime governance of production AI agents, separating a single reasoning-plane adjudication from coordinated enforcement across four infrastructure planes (Section 4).
\end{enumerate}

\begin{enumerate}
\item \textbf{Stop-anywhere mediation as a design property}, formalized over seven mediation points, with a taxonomy of six interruption primitives that generalizes allow and deny (Section 5).
\end{enumerate}

\begin{enumerate}
\item \textbf{A composite principal model with capability attenuation as a structural primitive}, with a linear-time adjudication algorithm and a capability-set lattice that admits static analysis (Section 6).
\end{enumerate}

\begin{enumerate}
\item \textbf{An audit substrate that produces reconstructible evidence}, with the property that partial evidence soundly bounds the system's behavior (Section 7).
\end{enumerate}

\begin{enumerate}
\item \textbf{Four correctness invariants} --- composed authority, mediation coverage, bounded composite authority, and evidence sufficiency --- argued structurally and stated precisely enough to be amenable to subsequent formal verification (Sections 4--7).
\end{enumerate}

\begin{enumerate}
\item \textbf{A demonstration of foreclosure} of seven production-agent threats across a canonical workflow and four further production use cases spanning financial services, healthcare, software engineering, and customer operations (Sections 3, 9), and \textbf{an evaluation framework} that treats safety and utility as joint objectives (Section 10).
\end{enumerate}

\subsection{Scope and roadmap}

The architecture governs \textit{action}, not \textit{model behavior}. It operates downstream of model alignment --- it constrains what an agent is permitted to do, not what the model is induced to say --- and upstream of infrastructure enforcement, which it directs but does not replace. This scoping is deliberate and is developed in the threat model (Section 3).

The remainder of the paper proceeds as follows. Section 2 situates the architecture in the lineage of security principles, capability theory, and modern authorization systems it extends. Section 3 specifies the threat model. Sections 4 through 7 develop the four primitives and state and argue for the four correctness invariants. Section 8 composes the primitives into a single reference architecture and traces a complete agent action through it. Section 9 demonstrates the architecture's foreclosure of the seven threats across a canonical workflow and four further production use cases. Section 10 proposes the evaluation framework. Section 11 discusses limitations, open problems, adversarial considerations, and the adoption path. Section 12 concludes.

\section{Background and Related Work}

The architecture this paper presents extends three established research traditions and responds to a fourth, emerging one. We organize the related work accordingly: the classical security principles the architecture applies to a new system class (Section 2.1); the object-capability theory from which the composite principal model descends (Section 2.2); the modern authorization systems the reasoning plane extends (Section 2.3); the zero-trust architectures whose trajectory the five-plane decomposition continues (Section 2.4); and the agentic-AI security literature the architecture is positioned within (Section 2.5). We close by situating the architecture relative to our own prior work on authorization in multi-agent systems (Section 2.6).

\subsection{Classical Security Principles}

The principles that govern the architecture are not new; their application to agentic systems is. Saltzer and Schroeder's foundational enumeration \cite{SaltzerSchroeder75} of the principles of secure system design --- least privilege, complete mediation, separation of privilege, fail-safe defaults, economy of mechanism, least common mechanism, psychological acceptability, and the work-factor and compromise-recording principles --- remains the canonical reference for what a secure system must do. The architecture is, in a precise sense, an application of these principles to a system class Saltzer and Schroeder could not have anticipated.

The correspondence is direct. \textit{Least privilege} is realized by the composite principal's capability attenuation (Section 6): an agent holds the minimum authority its delegation chain confers, never more. \textit{Complete mediation} is realized by stop-anywhere mediation (Section 5), extended from its classical scope (every access to every object is checked) to the agentic scope, in which every step of the agent's reasoning loop is mediated. \textit{Fail-safe defaults} are realized by the architecture's fail-closed failure semantics (Section 4.4): an action does not proceed unless every applicable plane affirmatively realizes its directive. \textit{Separation of privilege} is realized by the session-state predicates that forbid an agent from composing capabilities whose combination is dangerous (Section 6.4). And the \textit{compromise-recording} principle --- Saltzer and Schroeder's recognition that systems which cannot prevent compromise should at least record it reliably --- is realized by the audit substrate (Section 7), which we treat not as a fallback but as a first-class primitive.

What the architecture contributes over the classical principles is not a new principle but a new realization. Saltzer and Schroeder wrote for systems in which the principals were processes and users and the objects were memory and files. The agentic regime introduces principals that are composite and delegated, objects that are systems of record reached through connectors, and actions whose composition --- not whose individual occurrence --- is the threat. The architecture is the application of unchanged principles to this changed setting.

\subsection{Object-Capability Theory}

The composite principal model (Section 6) descends from object-capability theory. The capability model of access control --- in which authority is represented by unforgeable, transferable tokens that both designate a resource and confer the right to access it --- originates in the operating-systems literature \cite{DennisVanHorn66} and was developed into a unified theory of secure composition by Miller \cite{Miller06}. The central insight the architecture inherits is that \textit{authority should be attenuable}: the holder of a capability should be able to derive a weaker capability to delegate, but never a stronger one. This is the property the architecture elevates to a structural primitive in Definition 4.

The Confused Deputy problem \cite{Hardy88} is the canonical demonstration of what goes wrong without capability discipline: a program with authority is induced by a less-authorized caller to misuse that authority on the caller's behalf, because the program conflates the authority to act with the designation of what to act upon. The agentic regime reproduces the Confused Deputy at scale --- every agent that acts on behalf of a chain of delegators is a potential confused deputy --- and the composite principal model forecloses it structurally, by bounding the deputy's effective authority to the intersection of the chain (Section 6.2).

The closest prior art to the composite principal at the credential layer is the macaroon \cite{Birgisson14}: a bearer credential with contextual caveats that may only restrict, never expand, the bearer's authority. Macaroons pioneered attenuation-as-primitive in a deployable credential format, with chained caveats and cryptographic binding. The composite principal extends the macaroon insight from the credential layer to the runtime principal layer: where a macaroon is presented and verified at an access point, the composite principal is adjudicated against policy at the reasoning plane, carries its full delegation chain for policy to reason over, and integrates per-capability time-to-live as a structural property. We regard the macaroon as the architecture's most direct intellectual ancestor, and the composite principal as its extension to the multi-agent runtime regime.

\subsection{Modern Authorization Systems}

The reasoning plane (Section 4.2) is, in one view, an authorization system, and the architecture's relationship to existing authorization systems is one of extension rather than replacement.

Zanzibar \cite{Pang19}, Google's global authorization system, expresses authorization as a graph of relationships among users, groups, and objects, with computed usersets and caveats for conditional derivation. Zanzibar and its open-source descendants \cite{SpiceDB,Permify} are the state of the art for relationship-based access control, and they excel at the problem they were designed for: consistent, low-latency authorization over large, relatively static relationship graphs. The architecture positions Zanzibar-derived systems as a substrate for the \textit{base} capability sets of identifiable principals (Section 6.5.3), over which the composite principal model performs runtime delegation-chain adjudication. The two are complementary: Zanzibar answers ``what is principal $p$'s base authority over resource $r$?``; the composite principal model answers ``what authority does this delegation chain confer right now, given each delegator's base authority?''

Policy-as-code systems --- Open Policy Agent with the Rego language \cite{OPA}, and Amazon's Cedar \cite{Cedar} --- provide declarative, request-time evaluation of authorization policy. They are expressive, analyzable, and widely deployed. The architecture's reasoning plane requires capabilities these systems do not natively provide (Section 8.4): composite-principal quantification, stateful session predicates, the six interruption primitives as output types, and per-plane projection. We do not claim these systems are inadequate for their purpose; we claim that production-agent governance requires an evaluation model that extends them. The pragmatic path we identify (Section 8.4) treats an existing policy language as the substrate for the request-time portion of adjudication, wrapped by the stateful, multi-primitive, multi-plane logic the architecture adds --- a path that preserves the considerable investment in existing policy languages while supplying what they lack.

The architecture's stance toward these systems is consistent: each is the right tool for the problem it was designed for, and the architecture composes them rather than displacing them. This is not diplomatic hedging; it is the architecture's actual design. The reasoning plane is not a new authorization engine built from nothing --- it is a stateful, plan-aware, composite-principal-aware adjudicator that incorporates existing authorization engines for the portions of its decision they already evaluate well.

\subsection{Zero-Trust and the Relocation of the Trust Decision}

The five-plane decomposition (Section 4) continues a trajectory in enterprise security architecture: the progressive relocation of the trust decision closer to the action it governs.

The network-perimeter model located trust at the boundary between the corporate network and the outside: inside was trusted, outside was not. BeyondCorp \cite{WardBeyer14,BeyondCorp} relocated the trust decision from the network perimeter to an identity-aware proxy that adjudicates each request against the requesting user's identity and the requesting device's posture --- a two-plane model (identity and endpoint) with the reasoning function embedded in the proxy. The NIST zero-trust architecture \cite{ZeroTrustNIST} generalized this relocation across the enterprise: no implicit trust is granted on the basis of network location, and every access is adjudicated against policy.

The five-plane architecture continues this trajectory. Where BeyondCorp relocated the trust decision to the per-request boundary and zero-trust generalized it across the enterprise, the architecture relocates it to the per-step action of an agent --- adjudicated against the composite principal the agentic regime introduces, and realized across all four infrastructure planes (network, identity, endpoint, data) that an agent's action traverses, rather than the two BeyondCorp addressed. The architecture is, in this framing, the agentic generalization of zero-trust: it applies zero-trust's core commitment --- no implicit trust, every action adjudicated --- to a setting in which the actor is an agent, the principal is composite, and the unit of adjudication is the step rather than the request.

Service-mesh architectures \cite{Istio,Linkerd,Envoy} realize a related but narrower relocation: they adjudicate and enforce at the network plane, with mutual TLS and policy at the service boundary. The architecture treats the service mesh as a realization substrate for the network plane's directives (Section 4.3.1), consistent with its general stance of composing existing enforcement infrastructure rather than replacing it.

\subsection{Agentic AI Security}

The architecture is positioned within an emerging literature on the security of agentic AI systems, which has developed rapidly as tool-using agents have moved toward production.

Recent systematizations of agentic security argue for end-to-end systems security rather than model-only defenses. Work on systems-security foundations for agentic computing \cite{SoKAgentic} builds on established computer-security concepts to address the iterative, tool-invoking nature of agents; systematizations of the agentic attack surface \cite{SoKAttackSurface} and of Model Context Protocol security \cite{MCPSoK} enumerate the threat surface --- prompt injection, tool execution, external-server interaction, connector overreach --- that the agentic regime introduces. The architecture shares this systems-security framing and the conviction that model-layer defenses, while necessary, are insufficient: the security of an agentic system is a property of the deployed system, not of the model in isolation.

Indirect prompt injection \cite{Greshake23} --- in which adversarial content surfaced through an agent's tools manipulates its behavior --- is among the most studied agentic threats, and it appears as threat T1 in our enumeration (Section 3). Benchmarks such as AgentDojo \cite{Debenedetti24} provide a dynamic environment, with realistic email-, banking-, and travel-management tasks, against which injection attacks and defenses can be measured, and which we adopt as the anchor for the evaluation framework's attack corpus (Section 10.6). The architecture's response to indirect prompt injection is not content filtering but action governance: an injected instruction that influences the agent's plan is foreclosed at plan formation (MP1) if it requires authority the composite principal lacks, regardless of how persuasive the injected content is (Section 9.2). This reflects a general stance: the architecture defends against the \textit{effects} an injection seeks to produce, rather than attempting to detect the injection itself.

The architecture's capability-based stance is closely related to CaMeL \cite{Debenedetti25}, which defeats prompt injections by design: CaMeL extracts the control and data flows from the trusted query so that untrusted data can never influence program flow, and uses capabilities to prevent data exfiltration over unauthorized flows --- achieving provable foreclosure of an attack class at a modest utility cost (77\textbackslash{}\% versus 84\textbackslash{}\% task completion on AgentDojo). CaMeL operates at the interpreter layer of a single agent; the present architecture generalizes the capability-based commitment to the multi-agent runtime, with composite principals spanning delegation chains (Section 6), enforcement across four infrastructure planes (Section 4), and an audit substrate (Section 7). Our own work on operationalizing the CaMeL approach for enterprise deployment \cite{TallamCaMeL} informs the architecture's emphasis on deployability and on defenses that survive contact with production systems of record. The planner-executor separation, developed for capability rather than security reasons \cite{PlanAndAct,Yao22}, is repurposed by the architecture as a control surface: the separation of plan formation from execution provides a natural locus --- the reasoning plane between planner and executor --- at which intent can be adjudicated before action (Section 4.2, Section 5.2).

A growing body of work addresses authorization, identity, and governance specifically for agentic and multi-agent systems. The architecture's contribution within this body is to provide a \textit{reference architecture} --- a structural decomposition with formal correctness properties --- rather than a point defense against a specific threat or a specific deployment. Where much of the agentic-security literature addresses individual threats (a defense against injection, a mechanism for tool sandboxing, a method for output filtering), the architecture addresses the governance problem structurally, foreclosing classes of threat through the composition of general primitives (Section 9.9).

\subsection{Relationship to Our Prior Work}

The architecture builds directly on a sequence of our prior work on authorization and governance in agentic systems, and we state the relationship explicitly.

The composite principal model (Section 6) develops, into a formal runtime model, the treatment of authorization propagation in multi-agent systems presented in prior work on identity governance as infrastructure \cite{TallamAuthProp}. That work argued that authorization in multi-agent systems must propagate through delegation chains as a first-class infrastructural concern; the present architecture provides the formal principal model, attenuation primitive, and adjudication algorithm that realize the argument.

The audit substrate's reconstructability property (Section 7.3) builds on prior work benchmarking authorization-limited evidence in agentic systems \cite{TallamPartialEvid}. That work formalized and measured the reconstruction of agent behavior under partial evidence; the present architecture incorporates reconstructability as a structural requirement of the evidence substrate, with the prior benchmark serving as the evaluation instrument for the evidence-completeness metric (Section 10.2).

The architecture's broader framing (that assurance is a property of the deployed system rather than of the model, and that governance must address the full lifecycle of delegated agency) develops themes from our work on systems-engineering perspectives for frontier AI \cite{TallamSystems}, on governance under self-modification \cite{TallamLayeredMut}, and on security-by-design frameworks for autonomous systems \cite{TallamSecByDesign}. The present paper differs from this prior work in kind: where the prior work established framings, threat models, and principles, the present paper provides a reference architecture --- a concrete structural decomposition with stated correctness invariants and an evaluation framework --- that those framings motivate.

Two further works develop adjacent constructs: execution envelopes, a shared admission contract for bounding backend agent execution requests \cite{TallamExecEnvelopes}; and a companion treatment of identity legibility under distributed cognition, in submission. The present architecture is self-contained and does not depend on either.

The cumulative relationship is that the present paper is the architectural synthesis of a research program: the prior work established that authorization propagation, partial-evidence accountability, and systems-level assurance are the right problems; this paper composes their solutions into a single reference architecture for the runtime governance of production agents.

\section{Threat Model}

The reference architecture developed in Sections 4 through 8 is a response to a specific class of threats: those that arise when a tool-using AI agent operates against an enterprise's systems of record, connectors, and data, at production scale and under production service-level expectations. This section specifies the system under consideration, the adversary model, and the enumeration of threats the architecture is designed to address. The seven threats enumerated here are referenced throughout the paper --- by the mediation points of Section 5, the case study of Section 9, and the evaluation framework of Section 10 --- and we therefore fix them precisely.

We make one framing commitment at the outset. The threats we enumerate are \textit{systems-security} threats, not model-behavior threats. We are not concerned, in this paper, with whether the underlying model is aligned, whether it hallucinates, or whether it can be jailbroken in the narrow sense of producing disallowed text. We are concerned with what happens when an agent --- whatever its internal behavior --- is given the authority to read context, call tools, invoke connectors, and modify systems of record. The threats below are the threats of \textit{delegated action}, not of model output. This boundary is deliberate: it is the boundary at which runtime governance operates, downstream of model alignment and upstream of infrastructure enforcement.

\subsection{System Assumptions}

We consider a production agent system with the following components.

A \textbf{production agent} comprises a planner that forms intentions, an executor that carries them out, and a set of tools and connectors through which the agent reads context and effects change. The agent operates on behalf of an \textbf{originating principal} --- a human user, a service, or an upstream agent --- from which it has received a delegated authority profile. The agent may itself delegate subtasks to \textbf{downstream agents}, forming a delegation chain (Section 6).

The agent operates against \textbf{enterprise systems of record}: document stores, databases, communication systems, calendars, and the connectors that bridge them. These systems hold data subject to classification, residency, and retention constraints. The agent's tools include read operations (retrieval, search, summarization inputs) and write operations (sending messages, scheduling events, modifying records, creating documents).

The system runs at \textbf{enterprise service-level expectations}: agentic actions occur at timescales of roughly 100 milliseconds to several seconds, and the governance layer must operate within this budget (Section 8.5). The system serves multiple originating principals concurrently, with isolation requirements between them.

We assume the existence of the four infrastructure planes (network, identity, endpoint, data) as enforcement substrates with their standard primitives (Section 4.3), and the existence of an identity provider that can issue and validate credentials for principals including agent-class principals.

\subsection{Adversary Model}

We consider adversaries with varying access and capability.

\textbf{Malicious or compromised users.} An originating principal may be adversarial, or a legitimate principal's credentials may be compromised. Such an adversary seeks to induce the agent to take actions beyond the principal's authority, or to exfiltrate data the principal is not entitled to.

\textbf{Adversarial content sources.} Content the agent retrieves --- documents, web pages, emails, tool outputs --- may contain adversarial material crafted to manipulate the agent's behavior. The adversary controls the content but not the agent directly; the attack vector is the content surfaced to the agent through its tools.

\textbf{Compromised connectors and tools.} A connector or tool the agent invokes may be compromised, returning manipulated outputs or exceeding its stated function. The adversary controls the tool's behavior but not the reasoning plane.

\textbf{Malicious agents in multi-agent systems.} In a multi-agent workflow, one agent may be adversarial --- compromised, misconfigured, or operating under a manipulated delegation. Such an agent seeks to exploit its position in the delegation chain to exceed the authority the chain should confer.

\textbf{Insiders with elevated authority.} A principal with legitimate elevated authority may misuse it, or have it misused on their behalf through a confused-deputy pattern (Section 6.2).

We assume adversaries may read content, inject content into sources the agent retrieves, manipulate tool outputs, and attempt to replay or forge credentials. We assume adversaries cannot forge cryptographic signatures of principals whose keys they do not hold (Section 6.6), and cannot alter sealed evidence records without detection (Section 7.4). We assume the reasoning plane itself is within the trust boundary; its compromise is discussed as an adversarial consideration in Section 11, but is out of scope for the threats enumerated here.

\subsection{Threat Enumeration}

We enumerate seven threats. The enumeration is organized by the locus of the threat in the agent's operation --- what the agent reads, how it chains actions, what authority it carries, what it commits, and what evidence it produces. It is informed by, and broadly consistent with, the threat surfaces systematized in recent agentic-security work \cite{SoKAgentic,SoKAttackSurface,MCPSoK}, but it is organized for our purpose: each threat is positioned against the mediation points (Section 5.2) at which the architecture addresses it. For each threat we state the threat, its mechanism, and those mediation points. The detailed walkthrough of how the composed architecture forecloses each threat is the subject of the case studies in Section 9.

\textbf{T1. Indirect prompt injection through tool outputs.} Adversarial content surfaces in data the agent retrieves --- a document, an email, a web page, a tool response --- and manipulates the agent's plan, inducing it to take actions the originating principal did not intend \cite{Greshake23}. The mechanism is that the agent reasons over retrieved content as though it were trusted instruction; benchmarks of tool-using agents show current systems succumb to such injections in a substantial fraction of cases absent defenses \cite{Debenedetti24}. The architecture is positioned to address T1 at MP1 (plan formation, where a manipulated plan can be detected against the composite principal's authority) and MP2 (context retrieval, where the candidate set is constrained before adversarial content reaches the agent). Our stance follows capability-based defenses \cite{Debenedetti25} in foreclosing the injected \textit{effect} rather than attempting to detect the injecting \textit{content}.

\textbf{T2. Tool chain abuse.} A sequence of individually-permitted tool calls produces a composite effect that policy would forbid. The canonical instance is the read-then-exfiltrate chain: the agent reads a sensitive document (permitted) and then sends a message to an external recipient (permitted), where the sequence --- reading internal data and then transmitting externally in the same context --- is the exfiltration path. The mechanism is that per-call authorization cannot detect the violation, because each call is individually permitted. The architecture is positioned to address T2 at MP1 (where the plan as a whole, not the individual call, is evaluated) and through session-state predicates (Section 6.4) that forbid the chain even when each link is permitted.

\textbf{T3. Connector overreach.} An agent granted access to a resource through a connector accesses more than the task requires --- reading an entire mailbox when the task concerns one thread, or an entire document store when the task concerns one file. The mechanism is that connectors typically grant coarse-grained access (a scope, an OAuth grant) that exceeds the fine-grained access the task justifies; the Model Context Protocol ecosystem in particular has been analyzed as granting tool access at scopes substantially broader than individual tasks require \cite{MCPSoK}. The architecture is positioned to address T3 at MP2 (where the retrieval candidate set is constrained to the task-justified subset before retrieval) and MP3 (where the tool is bound under an attenuated capability set via the Narrow primitive).

\textbf{T4. Approval evasion.} An agent takes a high-consequence action without the human approval that policy requires, because the approval was bound to a specific action signature rather than to a capability boundary, and the agent reached the action through a path the approval rule did not anticipate. The mechanism is that procedural approval rules --- ``require approval to send external email'' --- can be evaded by an agent that reaches the consequential effect through a different action than the rule names. The architecture is positioned to address T4 by binding approval to capability boundaries rather than action signatures (Section 5.4.2, the Escalate primitive), so that any action requiring a capability in the approval-gated set triggers escalation regardless of the path by which the agent reached it.

\textbf{T5. Delegation chain exploitation.} A composite principal preserves more authority than any single hop in the delegation chain should confer, because the delegation model flattens the chain or fails to attenuate authority across it. The mechanism is that an intermediary agent --- legitimate or compromised --- exercises authority it should not hold, because the system evaluates against the acting principal's raw capabilities rather than the attenuated intersection of the chain. The architecture is positioned to address T5 structurally, through the composite principal model with capability attenuation as primitive (Section 6.2): the effective capability set is the intersection of the chain's capability sets, so no agent in the chain can exercise authority outside that intersection.

\textbf{T6. Audit opacity.} After an agent has operated, the system cannot reconstruct what it did, under whose authority, against what data, with what outcome --- because the audit function emitted unstructured logs rather than structured evidence, or because the evidence is incomplete, untrustworthy, or unavailable. The mechanism is that logging-as-audit (Section 7.1) produces records insufficient for reconstruction. The architecture is positioned to address T6 through the audit substrate (Section 7): every decision produces a structured, complete, tamper-evident, reconstructible evidence record.

\textbf{T7. Workflow integrity loss.} An agent modifies the business process itself --- not merely accessing data, but changing the state of systems of record in ways that transform the workflow the originating principal authorized into a different workflow. The mechanism is that an agent with write authority can effect changes whose composition alters the process, even when each individual change is within authority. This is the threat that most distinguishes the agentic regime from the data-at-rest regime: the risk is not that data leaves a boundary, but that the agent's actions transform the process. The architecture is positioned to address T7 at MP1 (where the plan's aggregate effect on systems of record is evaluated), MP5 (where each state-changing commit is mediated), and through the Rollback primitive (Section 5.4.6, where a process-altering effect can be compensated).

We summarize the seven threats and their primary mediation points in the following table.

\begin{table}[ht]
\centering
\small
\caption{The seven production-agent threats and the primary mediation points at which the architecture addresses each.}
\label{tab:threats}
\begin{tabularx}{\textwidth}{lLL}
\toprule
Threat & Locus & Primary mediation points \\
\midrule
T1 Indirect prompt injection & What the agent reads & MP1, MP2 \\
T2 Tool chain abuse & How the agent chains actions & MP1, session-state \\
T3 Connector overreach & What the agent accesses & MP2, MP3 \\
T4 Approval evasion & What requires human sign-off & MP1, Escalate \\
T5 Delegation chain exploitation & What authority the agent carries & Composite principal (structural) \\
T6 Audit opacity & What evidence the agent produces & MP7, audit substrate \\
T7 Workflow integrity loss & What the agent commits & MP1, MP5, Rollback \\
\bottomrule
\end{tabularx}
\end{table}

The seven threats are not exhaustive of all possible attacks on agentic systems; they are the threats that \textit{delegated action} introduces and that \textit{runtime governance} is positioned to address. We argue (Section 4.5) that each is addressable at one or more of the five planes, and we demonstrate (Section 9) the foreclosure of each on a concrete production workflow.

\subsection{Out of Scope}

We delimit the threat model by stating what it does not cover.

\textbf{Model integrity and alignment.} Whether the underlying model is aligned, produces accurate outputs, or can be induced to generate disallowed text is upstream of the architecture. Runtime governance constrains what the agent is permitted to \textit{do}, not what the model is induced to \textit{say}. A model that proposes a forbidden action is handled by the architecture (the action is foreclosed); a model that produces an inaccurate summary is a model-quality concern the architecture does not address.

\textbf{Hardware and platform attestation.} The integrity of the compute substrate on which the agent and the reasoning plane run is the province of confidential computing and platform attestation, which the endpoint plane (Section 4.3.3) consumes but does not itself provide. We assume the endpoint plane can attest runtime posture; how that attestation is rooted is out of scope.

\textbf{Side-channel and timing attacks.} Attacks that extract information through timing, power, or other side channels are the province of classical systems security and are out of scope for the action-governance threats considered here.

\textbf{Denial of service.} Attacks that seek to exhaust the system's resources rather than to exceed authority are out of scope; the architecture's rate-limiting at the network plane (Section 4.3.1) provides partial mitigation, but DoS is not among the threats the architecture is designed to address.

These exclusions sharpen the architecture's claim. The reference architecture is a response to the threats of delegated action --- the threats that arise specifically because an agent has been given authority to act on an enterprise's behalf. It is not a general-purpose security system, and it does not displace the model-alignment, platform-attestation, and classical-systems-security layers above and below it. It occupies the layer between them: the runtime governance of what a production agent is permitted to do.

\section{The Five-Plane Reference Architecture}

\begin{figure}[t]
\centering
\begin{tikzpicture}[
  >=Latex,
  font=\small,
  agent/.style   = {draw, rounded corners=3pt, fill=black!4, align=center,
                    inner sep=6pt, minimum width=3.4cm, minimum height=10mm},
  reason/.style  = {draw, very thick, rounded corners=3pt, fill=black!10, align=center,
                    inner sep=6pt, minimum width=6.6cm, minimum height=13mm},
  plane/.style   = {draw, rounded corners=2pt, fill=black!3, align=center,
                    inner sep=5pt, minimum width=2.5cm, minimum height=11mm},
  store/.style   = {draw, dashed, rounded corners=2pt, align=center,
                    inner sep=5pt, minimum width=3.0cm, minimum height=10mm},
  audit/.style   = {draw, rounded corners=2pt, fill=black!6, align=center,
                    inner sep=5pt, minimum width=9.0cm, minimum height=9mm},
  arr/.style     = {-Latex, thick},
  darr/.style    = {Latex-Latex, thick}
]

\node[agent] (agent) {\textbf{Agent runtime}\\[1pt]
  \scriptsize planner \,\textbullet\, executor \,\textbullet\, tools};

\node[reason, below=10mm of agent] (reason)
  {\textbf{Reasoning plane} — adjudicator\\[1pt]
   \scriptsize evaluates $\langle \Pi, \pi, s, \mathcal{P}\rangle$, emits decision projection $D(a)$};

\node[store, left=8mm of reason] (cps)
  {\textbf{Composite}\\\textbf{principal} $\Pi$\\\scriptsize attenuated, signed};
\node[store, right=8mm of reason] (pol)
  {\textbf{Policy}\\\textbf{store} $\mathcal{P}$\\\scriptsize versioned};

\node[plane, below=12mm of reason.south, xshift=-3.9cm] (net)
  {\textbf{Network}\\\scriptsize mTLS, segmentation};
\node[plane, right=4mm of net] (id)
  {\textbf{Identity}\\\scriptsize short-lived creds};
\node[plane, right=4mm of id] (ep)
  {\textbf{Endpoint}\\\scriptsize posture, attest.};
\node[plane, right=4mm of ep] (data)
  {\textbf{Data}\\\scriptsize classify, pre-retrieval};

\node[audit, below=10mm of id.south, xshift=1.3cm] (audit)
  {\textbf{Audit substrate} — composes evidence record $e(a)$ \;\scriptsize(structured, tamper-evident, reconstructible) $\rightarrow$ SIEM / OTel};

\draw[darr] (agent) -- node[right, font=\scriptsize] {7 mediation points} (reason);
\draw[darr] (reason) -- (cps);
\draw[darr] (reason) -- (pol);
\draw[arr] (reason.south) -- ++(0,-0.4) -| (net.north);
\draw[arr] (reason.south) -- ++(0,-0.4) -| (id.north);
\draw[arr] (reason.south) -- ++(0,-0.4) -| (ep.north);
\draw[arr] (reason.south) -- ++(0,-0.4) -| (data.north)
   node[pos=0.25, above, font=\scriptsize] {decision projection $\langle P_{\mathrm{net}},P_{\mathrm{id}},P_{\mathrm{ep}},P_{\mathrm{data}}\rangle$};
\draw[arr] (net.south)  |- (audit.west);
\draw[arr] (id.south)   -- (audit.north west);
\draw[arr] (ep.south)   -- (audit.north east);
\draw[arr] (data.south) |- (audit.east);
\draw[arr, dashed] (reason.east) to[out=0,in=90] (audit.north east);

\end{tikzpicture}
\caption{The five-plane reference architecture.}
\label{fig:five-plane}
\end{figure}
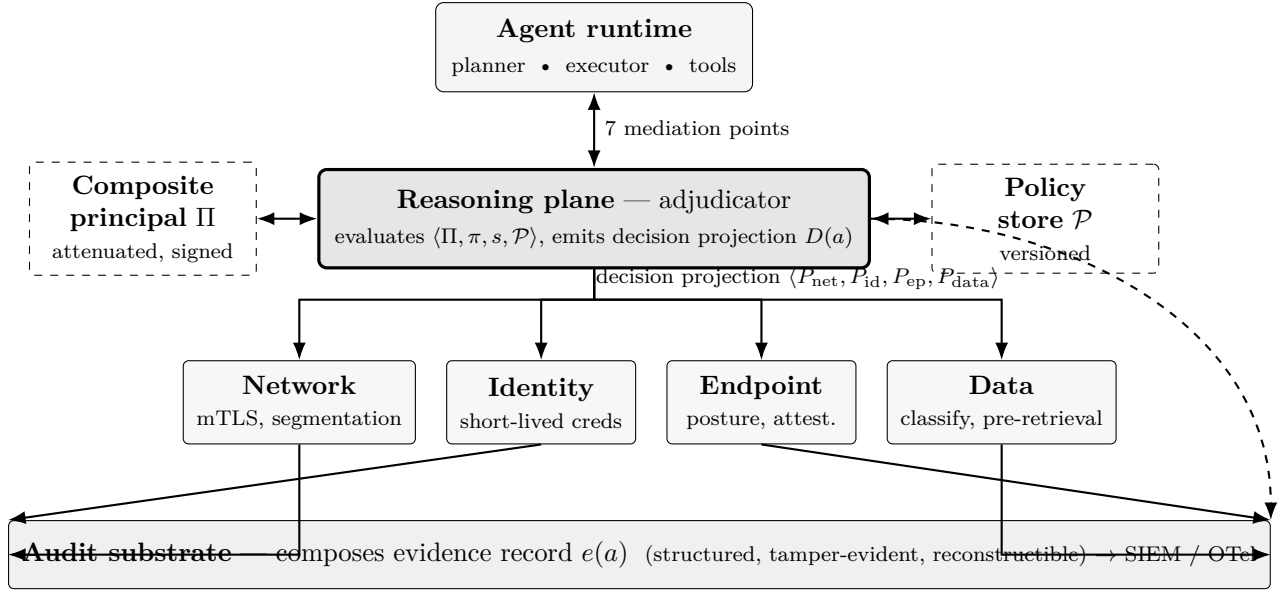

A production agent decision touches five planes simultaneously: the \textit{reasoning plane}, where intent is formed and authority is adjudicated; and four \textit{infrastructure planes} --- network, identity, endpoint, and data --- where the decision is realized as observable action against the enterprise's existing fabric. Prior architectures place policy enforcement at one or more of the infrastructure planes individually, with each plane operating under a distinct policy engine, ownership model, and audit substrate. The reference architecture we present in this paper unifies these planes under a single adjudication: the reasoning plane produces a structured \textit{decision projection} that the four infrastructure planes realize in parallel. The decision is made once, against the full agentic context, and enforced across the planes that an agent's action actually traverses.

This section defines each plane, specifies the contract between them, and argues that the five-plane decomposition is both necessary and sufficient for governing production AI agents at enterprise scale. We compare the resulting architecture with three prior reference architectures --- BeyondCorp \cite{WardBeyer14}, service-mesh deployments \cite{Istio,Linkerd}, and stateless authorization systems \cite{OPA,Cedar} --- and show that each is a degenerate case of the five-plane composition: necessary at its layer, insufficient when an agentic action traverses every layer in a single decision.

\subsection{Overview}

The architectural commitment of this paper is the following:

\begin{claim}[Adjudication-Enforcement Separation]
In a production agent system, the authority to act is decided at a single adjudication point --- the reasoning plane --- that has visibility into intent, composite principal, and accumulated session state; and is realized through coordinated enforcement at four infrastructure planes that already exist in the enterprise. Policy is composed, not federated.
\end{claim}

This commitment differs structurally from the prevailing decomposition, in which each infrastructure plane carries its own policy engine and audit pipeline. Under federated enforcement, a single agentic action --- say, an autonomous schedule-and-notify workflow --- may produce four independent policy decisions: one at the network firewall, one at the identity provider, one at the endpoint posture check, and one at the data classification gate. Each decision is made against partial context. Each emits its own audit record. None of them sees the full plan that the agent has formed, or the chain of delegations through which the agent acquired the authority to attempt it. The composition of these independent decisions is, in the general case, unsound: an action may be authorized at every plane individually and yet violate the composite principal's actual capability set.

The reference architecture forecloses this failure mode by structural separation. The reasoning plane is the only locus of policy decision. The infrastructure planes are no longer independent decision-makers; they are \textit{enforcement substrates} that realize a decision made elsewhere. The audit substrate is correspondingly unified: a single evidence record, produced by the reasoning plane and annotated by each enforcement plane with its realization status, replaces four independent log streams. Figure\textasciitilde{}\ref{fig:five-plane} summarizes the composition: one adjudication at the reasoning plane fans out to the four enforcement planes and fans back in to a single evidence record.

The remainder of this section formalizes the planes and their interaction.

\subsection{The Reasoning Plane}

The reasoning plane is the adjudication layer. It is the only plane that has full visibility into the agentic decision context, and it is the only plane authorized to issue a decision.

\textbf{Definition 1 (Reasoning Plane).} \textit{The reasoning plane is a stateful policy adjudicator that, for each proposed agent action $a$, evaluates the tuple $\langle p, \pi, s, h \rangle$ --- where $p$ is the composite principal, $\pi$ is the agent's current plan, $s$ is the session state, and $h$ is the accumulated decision history --- and produces a decision projection $D(a) = \langle o, P_{\mathrm{net}}, P_{\mathrm{id}}, P_{\mathrm{ep}}, P_{\mathrm{data}}, e \rangle$, where $o$ is one of the six interruption primitives (Section 5), each $P_x$ is the per-plane enforcement directive for plane $x$, and $e$ is the structured evidence record.}

The reasoning plane is the only plane in the architecture that exists \textit{because of} the shift to agentic systems. Networks, identity providers, endpoints, and data systems pre-date production AI deployments by decades, and each has well-developed enforcement primitives. What did not previously exist --- and what existing policy engines do not provide --- is a layer that adjudicates \textit{intent before action} across the composite principal context that agentic systems introduce.

We make three claims about the reasoning plane's necessity.

First, an agent's plan can violate policy before any action manifests at the infrastructure planes. A workflow in which an agent retrieves a sensitive document, summarizes it, and then sends the summary to an external recipient may violate policy at the composition level even though each individual action is permitted in isolation. Enforcement-only architectures --- those without a reasoning plane --- detect such violations only after the violating action has fired, at which point compensation is required. The reasoning plane allows the decision to be made \textit{before} the plan executes.

Second, composite principals cannot be evaluated at any single infrastructure plane. The identity plane sees the acting principal, not the chain through which authority was delegated. The data plane sees the resource, not the policy clause that governs whether the composite principal may touch it. Composite-principal evaluation requires a plane that holds the full chain --- which is exactly the reasoning plane's role.

Third, session-state-dependent decisions cannot be expressed in stateless evaluators. Decisions of the form ``an agent that has already retrieved customer-A records may not now retrieve customer-B records in the same session'' require the engine to carry session history. OPA-class evaluators are by design stateless \cite{OPAdesign}; encoding session state in their input creates correctness and performance pathologies. The reasoning plane is stateful by construction.

The reasoning plane's output is not a Boolean; it is a \textit{decision projection} --- a structured object that specifies, for each of the four infrastructure planes, what realization is required. We treat this projection as the architecture's canonical inter-plane contract, and develop its structure in Section 4.4.

\subsection{The Four Infrastructure Planes}

We adopt the conventional decomposition of enterprise security infrastructure into four planes \cite{BeyondCorp,ZeroTrustNIST}: network, identity, endpoint, and data. Each plane has well-developed enforcement primitives, well-understood ownership models, and a substantial body of operational practice. The architecture's claim is not that these planes are new; it is that they are \textit{insufficient as independent policy authorities} for production-agent governance, and must be reorganized as enforcement substrates beneath the reasoning plane.

\subsubsection{The Network Plane}

The network plane enforces decisions about which endpoints may communicate, under what cryptographic posture, and at what rate. Its primitives include firewalls, microsegmentation, service-mesh policies, mutual TLS, and rate-limiting. Its enforcement latency is sub-millisecond; its decisions are made at line rate against compiled rule sets. Its operational ownership in most enterprises sits with security engineering or network operations.

For production-agent governance, the network plane realizes decisions of the form: \textit{this composite principal may invoke this service endpoint from this origin under this rate budget}. The plane's enforcement is well-suited to per-action evaluation at the network boundary, but it has no visibility into the agentic context (plan, composite principal, session state) that determines whether the action should be permitted in the first place. Under the reference architecture, the network plane receives a directive $P_{\mathrm{net}}$ that has already been adjudicated; the network plane's task is to realize the directive, not to evaluate authority.

\subsubsection{The Identity Plane}

The identity plane enforces decisions about who is acting and under what scope. Its primitives include identity providers (Keycloak, Okta, Azure AD), token validation, scope assertion, federation, and short-lived credentialing. Its enforcement latency is tens of milliseconds; its decisions are made against signed assertions that carry claims about the principal.

For production-agent governance, the identity plane realizes decisions about the \textit{acting principal} --- the immediate agent that is invoking a tool --- but does not natively reason about the \textit{composite principal}, the full chain through which authority arrived at the acting agent. OAuth on-behalf-of flows \cite{OAuthOBO} flatten the chain to the last hop; SAML assertion delegation has analogous limitations. Under the reference architecture, the identity plane receives a directive $P_{\mathrm{id}}$ that specifies the attenuated capability set the composite principal carries for this action; the plane realizes the directive by binding short-lived credentials that reflect the attenuation, rather than the acting principal's raw capabilities.

\subsubsection{The Endpoint Plane}

The endpoint plane enforces decisions about the runtime environment in which the action executes. Its primitives include device-posture checks, runtime attestation, configuration boundaries, and endpoint protection. Its enforcement latency is variable; some decisions are precomputed and cached, others require live attestation. Its operational ownership in most enterprises is shared between IT and security.

For production-agent governance, the endpoint plane realizes decisions of the form: \textit{this action may execute only on a runtime that satisfies this posture predicate}. The plane is the architecture's analog of the device-posture layer in BeyondCorp \cite{BeyondCorp}; what it adds, for agentic systems, is awareness that the ``endpoint'' may be a serverless function, a container, an agent runtime, or a privileged tool host. Under the reference architecture, the endpoint plane receives a directive $P_{\mathrm{ep}}$ that specifies the posture envelope, and refuses the action if the envelope is not satisfied.

\subsubsection{The Data Plane}

The data plane enforces decisions about which data the action may touch, at what granularity, and under what lineage and residency constraints. Its primitives include data classification systems, lineage trackers, residency controllers, and tokenization or encryption gateways. Its enforcement latency is the most variable of the four planes; per-record evaluation against fine-grained classifications is expensive enough that amortization or precomputation is often required.

For production-agent governance, the data plane is the plane at which the agent's actual contact with enterprise content occurs. Under the federated model, the data plane carries its own policy engine (commonly a DLP system) that decides what may leave a boundary. We argue that this configuration is precisely what fails under production agents: the risky thing is no longer that data crosses a boundary, but that the agent makes a sequence of decisions about which data to consult, summarize, transform, or propagate. A data-plane DLP engine cannot evaluate the agent's plan; it can only inspect content in transit. Under the reference architecture, the data plane receives a directive $P_{\mathrm{data}}$ that constrains the candidate set the agent may retrieve, \textit{before} retrieval occurs --- converting the data plane from a post-hoc filter to a pre-retrieval gate \cite{TallamAuthProp,TallamPartialEvid}.

\subsection{Cross-Plane Composition}

The reasoning plane's decision projection $D(a)$ specifies four per-plane directives that must be realized in coordination. We define the composition protocol that connects them.

\textbf{Fan-out.} Upon producing $D(a)$, the reasoning plane dispatches each directive $P_x$ to its corresponding enforcement plane. The dispatch is asynchronous; the reasoning plane does not block on any individual plane's response.

\textbf{Plane-local realization.} Each enforcement plane realizes its directive using its native primitives. The network plane installs or updates filter rules; the identity plane issues short-lived credentials reflecting the attenuated capability set; the endpoint plane checks posture and admits the action; the data plane constrains the retrieval candidate set. Each plane emits a realization-status record indicating success, failure, or partial realization.

\textbf{Fan-in.} The audit substrate (Section 7) collects the realization-status records from each plane and composes them with the reasoning plane's original decision projection into a single evidence record. This composed record is the canonical audit artifact for the action.

The composition protocol must address three coordination challenges that arise from the planes' heterogeneous latency, ownership, and enforcement models.

\textit{Latency reconciliation.} The four planes operate at different enforcement timescales: network at sub-millisecond, identity at tens of milliseconds, endpoint at variable latency including human-attested checks, data at potentially seconds for fine-grained classification. The reasoning plane cannot block on all of them synchronously without imposing an unacceptable per-action latency tax. The architecture distinguishes \textit{synchronous planes} --- those whose realization status gates the action --- from \textit{compensating planes} --- those whose realization is asynchronous and may trigger rollback (Section 5) if it fails. The mapping of plane to mode is policy-dependent; sensitive actions may require all four planes synchronous, while low-stakes actions may tolerate compensating data-plane realization.

\textit{Ownership reconciliation.} The four planes are owned by different organizational functions in most enterprises. The architecture does not require organizational unification; it requires that each plane accept a \textit{directive contract} from the reasoning plane and emit a realization-status record back to it. The planes' internal mechanisms --- what rule language they use, what platform they run on, who operates them --- are unchanged. What changes is that the planes no longer make policy decisions independently; they realize decisions made at the reasoning plane.

\textit{Failure semantics.} If an enforcement plane cannot realize its directive --- because the plane is unavailable, because the directive is malformed, or because the plane's own constraints conflict with the directive --- the action does not proceed. The architecture's default failure mode is \textit{fail-closed}: if any synchronous plane reports realization failure, the reasoning plane invokes its rollback primitive (Section 5.4.6) and emits a failure-mode evidence record. The architecture does not permit the action to proceed under partial realization; the audit substrate must be able to assert that every realized action was realized at every applicable plane.

We summarize the composition protocol in the following invariant:

\begin{invariant}[Composed Authority]
No action realized through the architecture has been authorized by any single plane in isolation. Every realized action carries an evidence record demonstrating that (a) the reasoning plane adjudicated it, (b) each applicable enforcement plane realized its directive, and (c) the realizations are jointly consistent with the reasoning plane's original decision projection.
\end{invariant}

This invariant is the architecture's central correctness property. It is what makes the five-plane composition sound where federated enforcement is not.

\subsection{Why Five and Only Five}

The choice of five planes is not arbitrary, but neither is it obvious. We argue that the decomposition is both \textit{necessary} --- fewer planes admit known failure modes --- and \textit{sufficient} --- additional planes either over-specify the architecture or conflate enforcement with telemetry.

\subsubsection{Necessity}

Each of the five planes addresses a class of threat that the others cannot.

\begin{itemize}
\item \textbf{Without a reasoning plane}, no plane has visibility into the agent's plan, composite principal, or session state. Plan-level violations cannot be detected before realization; composite-principal attenuation cannot be enforced. The system reduces to federated enforcement, which we argued above is unsound for composite principals.
\item \textbf{Without a network plane}, the architecture cannot enforce communication-level constraints --- which endpoints the agent may reach, under what cryptographic posture, at what rate. Adversarial connectors that exfiltrate via network calls are not foreclosed.
\item \textbf{Without an identity plane}, the architecture cannot issue credentials that reflect the attenuated capability set the reasoning plane has computed. The acting principal continues to carry its raw authority, and identity-bearing downstream systems cannot make consistent authorization decisions.
\item \textbf{Without an endpoint plane}, the architecture cannot constrain the runtime environment in which the action executes. Compromised hosts, missing attestation, and configuration drift remain unaddressed.
\item \textbf{Without a data plane}, the architecture cannot apply pre-retrieval constraints to the agent's contact with enterprise content. The agent retrieves more than the reasoning plane intended, and output filters become the only line of defense --- which, as we argued in Section 3, is too late for production-agent risk.
\end{itemize}

Each plane addresses a class of threat with no substitute. The composition is necessary.

\subsubsection{Sufficiency}

We considered three candidate additional planes and argue that each is more parsimoniously expressed within the existing five.

\begin{itemize}
\item \textbf{An observability plane.} Telemetry and metrics are essential to operating the architecture, but they do not enforce; they observe. Observability is a property of the audit substrate (Section 7) and a downstream consumer of evidence records, not a separate plane.
\item \textbf{A compliance plane.} Compliance requirements (residency, retention, attestation, evidence) are policy inputs to the reasoning plane, not a separate enforcement layer. A compliance plane would duplicate the reasoning plane's adjudication.
\item \textbf{A model plane.} Decisions about model behavior, alignment, and refusal are upstream of the architecture; they constrain what the agent can propose, not what the system permits the agent to do. The boundary between model behavior and runtime governance is the architecture's input boundary, not an additional plane.
\end{itemize}

The five-plane decomposition is sufficient: the classes of threat enumerated in Section 3 are each addressable at one or more of the five planes, and no class requires a sixth.

\subsection{Comparison with Prior Reference Architectures}

The five-plane architecture is best understood as the agentic extension of three prior reference architectures, each of which is recovered as a degenerate case.

\textbf{BeyondCorp} \cite{WardBeyer14,BeyondCorp} introduced the identity-aware proxy: the trust decision moves from the network perimeter to a per-request adjudication that combines identity and device posture. BeyondCorp is a two-plane architecture (identity + endpoint), with the reasoning function implicit in the proxy. The five-plane architecture preserves BeyondCorp's structural insight --- that trust decisions belong at the request boundary --- and extends it to a per-step agent action boundary, with explicit composition across all four infrastructure planes and an explicit reasoning plane above.

\textbf{Service mesh} architectures \cite{Istio,Linkerd,Envoy} place enforcement at the network plane, with authorization expressed in mesh-native policy languages. They are single-plane architectures: every decision is a network decision. They have no native model for composite principals, no native model for plan-level evaluation, and no native model for data- or endpoint-plane realization. The five-plane architecture is consistent with service-mesh enforcement at the network plane, and treats the mesh as a realization substrate for $P_{\mathrm{net}}$ directives.

\textbf{Stateless authorization systems} \cite{OPA,Cedar,Pang19} place enforcement at the application boundary, evaluating request-time decisions against compiled policy. They are reasoning-adjacent: they perform adjudication, but on a per-request basis, against an atomic principal, with no session state and no composite principal model. The five-plane architecture treats stateless authorization systems as a candidate substrate for the reasoning plane's decision evaluation --- extended with composite principal evaluation, session state, and the six interruption primitives (Section 5). The relationship is one of strict extension: Rego-style and Cedar-style policy expressions remain valid as inputs to the reasoning plane, augmented with the agentic constructs the underlying language must be extended to express.

The five-plane architecture is not, therefore, a replacement for prior reference architectures. It is the composition under which each prior architecture's contribution is preserved, while the gap each leaves --- the absence of a reasoning plane that adjudicates intent across composite principal context, projected onto coordinated enforcement at the planes the agent's action actually traverses --- is filled.

\section{Stop-Anywhere Mediation as a Design Property}

Section 4 established the architectural composition under which a single adjudication at the reasoning plane is projected onto coordinated enforcement at four infrastructure planes. The composition is structurally sound, but it does not yet specify what the reasoning plane \textit{can do} when an agentic action is presented to it. Traditional policy evaluators answer this question with a Boolean: allow or deny. We argue in this section that Boolean evaluation is insufficient for production-agent governance, and that the reasoning plane must instead carry a richer structured-output vocabulary --- what we call the \textit{six interruption primitives} --- that generalizes allow/deny and treats the agent's execution pipeline as a sequence of \textit{seven mediation points}, each of which the engine must be able to observe and, if policy requires, interrupt.

We define the resulting design property --- \textit{stop-anywhere mediation} --- formally, develop the seven mediation points and the six interruption primitives, specify the state preservation requirements at each mediation point, and address the hardest implementation problem the architecture creates: clean rollback in the presence of side-effecting agent actions. We close by formalizing the claim that traditional allow/deny systems are the degenerate case of a stop-anywhere mediator restricted to two primitives.

\subsection{Definition}

We define stop-anywhere mediation as a design property of a runtime governance system.

\begin{definition}[Stop-Anywhere Mediation]
A runtime governance system $G$ exhibits stop-anywhere mediation iff (a) $G$ is present at every meaningful point in the agent's execution pipeline --- what we call the seven mediation points (Section 5.2) --- and (b) at each mediation point, $G$'s output vocabulary is the set of six interruption primitives (Section 5.4), of which allow-and-deny are the two degenerate cases.
\end{definition}

The definition is structural rather than behavioral: it specifies what the system can do, not what it must do. A stop-anywhere mediator may pass through every mediation point without intervening; what makes it stop-anywhere is that the capacity to intervene exists at every point, and the vocabulary of intervention is rich enough to express the policy decisions a production-agent system requires.

We distinguish stop-anywhere mediation from two related properties. \textit{Complete mediation} in the classical Saltzer-Schroeder sense \cite{SaltzerSchroeder75} requires that every access to every object be checked; stop-anywhere mediation extends this from access to \textit{action sequence}, and from object to \textit{the agent's reasoning loop itself}. \textit{Total enforcement} requires that every detected violation be blocked; stop-anywhere mediation does not --- it requires that the system can \textit{intervene}, where intervention is one of six structured outcomes, only one of which is blocking. The distinction is important: it is what makes the architecture usable in production, where indiscriminate blocking destroys agent utility and indiscriminate allowance destroys agent safety.

\subsection{The Seven Mediation Points}

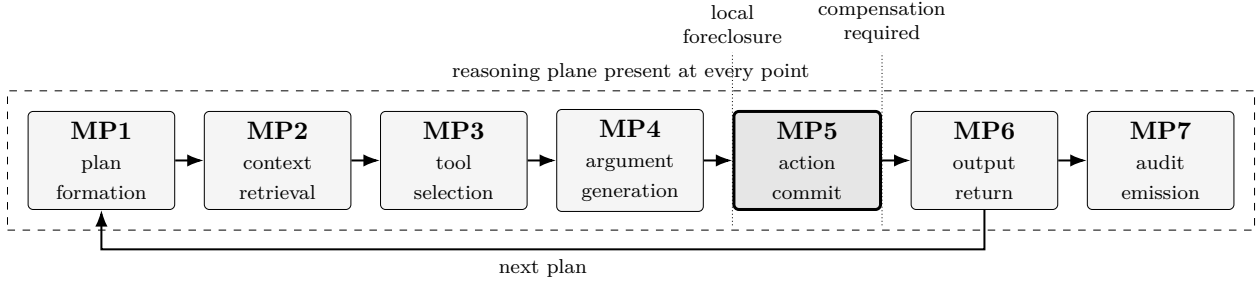
\begin{figure}[t]
\centering
\resizebox{\textwidth}{!}{%
\begin{tikzpicture}[
  >=Latex,
  font=\small,
  mp/.style   = {draw, rounded corners=2pt, fill=black!4, align=center,
                 inner sep=4pt, minimum width=2.05cm, minimum height=12mm},
  commit/.style = {draw, very thick, rounded corners=2pt, fill=black!10, align=center,
                 inner sep=4pt, minimum width=2.05cm, minimum height=12mm},
  arr/.style  = {-Latex, thick}
]
\node[mp]     (mp1) {\textbf{MP1}\\\scriptsize plan\\\scriptsize formation};
\node[mp, right=4mm of mp1]     (mp2) {\textbf{MP2}\\\scriptsize context\\\scriptsize retrieval};
\node[mp, right=4mm of mp2]     (mp3) {\textbf{MP3}\\\scriptsize tool\\\scriptsize selection};
\node[mp, right=4mm of mp3]     (mp4) {\textbf{MP4}\\\scriptsize argument\\\scriptsize generation};
\node[commit, right=4mm of mp4] (mp5) {\textbf{MP5}\\\scriptsize action\\\scriptsize commit};
\node[mp, right=4mm of mp5]     (mp6) {\textbf{MP6}\\\scriptsize output\\\scriptsize return};
\node[mp, right=4mm of mp6]     (mp7) {\textbf{MP7}\\\scriptsize audit\\\scriptsize emission};

\foreach \a/\b in {mp1/mp2,mp2/mp3,mp3/mp4,mp4/mp5,mp5/mp6,mp6/mp7}
  \draw[arr] (\a) -- (\b);

\draw[arr] (mp6.south) -- ++(0,-0.55) -| node[pos=0.25,below,font=\scriptsize] {next plan} (mp1.south);

\node[draw, dashed, fit=(mp1)(mp7), inner sep=8pt, label=above:{\scriptsize reasoning plane present at every point}] {};

\draw[densely dotted] ([yshift=-2mm]mp5.south west) -- ([yshift=8mm]mp5.north west)
  node[above,font=\scriptsize,align=center] {local\\foreclosure};
\draw[densely dotted] ([yshift=-2mm]mp5.south east) -- ([yshift=8mm]mp5.north east)
  node[above,font=\scriptsize,align=center] {compensation\\required};
\end{tikzpicture}%
}
\caption{The seven mediation points along the agent execution loop.}
\label{fig:mediation-points}
\end{figure}

A production agent executes a loop that, abstracted across runtime architectures (single-shot, agentic, multi-agent, planner-executor), exposes seven distinguishable points at which a runtime governance system may observe and intervene. We treat the seven-point decomposition as a \textit{structural property of agentic execution}, not as an implementation choice: regardless of whether the runtime is built on a planner-executor split \cite{PlanAndAct}, a single-loop ReAct-style agent \cite{Yao22}, or a multi-agent orchestrator, the seven points are present, though they may be more or less explicit in the runtime's API.

We enumerate the seven points in execution order; Figure\textasciitilde{}\ref{fig:mediation-points} shows them along the agent's execution loop, with the action-commit boundary (MP5) marked as the point past which foreclosure requires compensation.

\textbf{MP1. Plan formation.} The point at which the agent commits to a plan --- a structured intention about what to do next, expressed as a sequence of intended steps. Plan formation occurs explicitly in planner-executor architectures and implicitly in single-loop agents (where the plan is the next-token sequence the model is about to generate). Mediation at MP1 evaluates the \textit{plan as a whole}: its composition, its accumulated side effects, its consistency with the composite principal's authority. MP1 is the architecture's earliest opportunity to detect plan-level threats --- particularly threats 1, 2, and 6 from Section 3 (indirect prompt injection, tool chain abuse, workflow integrity loss).

\textbf{MP2. Context retrieval.} The point at which the agent retrieves external content --- documents, vector hits, search results, tool outputs that contain text the agent will subsequently reason over. Mediation at MP2 constrains the \textit{candidate set} the agent may consult, before the content reaches the reasoning loop. This is the architecture's pre-retrieval gate: the data plane realizes the directive $P_{\mathrm{data}}$ from the decision projection (Section 4.4) at MP2.

\textbf{MP3. Tool selection.} The point at which the agent commits to invoking a specific tool. Mediation at MP3 evaluates \textit{whether this tool is available to this composite principal at this stage of the plan}. MP3 is structurally distinct from MP4 (argument generation): the agent may be authorized to call a tool in principle while being unauthorized to call it with the particular arguments it is about to generate.

\textbf{MP4. Argument generation.} The point at which the agent has generated the arguments for the tool call but has not yet committed them. Mediation at MP4 evaluates \textit{the proposed argument set} --- its values, its types, its cross-references to other state. MP4 is the architecture's argument-rewriting opportunity: the Modify primitive (Section 5.4.4) is realized here.

\textbf{MP5. Action commit.} The point at which the agent's tool call crosses the system boundary --- the moment the call becomes externally observable, side-effecting, or otherwise difficult to compensate. Mediation at MP5 is the \textit{last point at which the action can be foreclosed without compensation}. Prior to MP5, blocking is a local operation; after MP5, blocking requires the rollback primitive.

\textbf{MP6. Output return.} The point at which the tool's response returns to the agent. Mediation at MP6 evaluates the \textit{response} --- its content, its provenance, its propagation properties. MP6 is the last point at which the system can intervene before the agent begins a new plan formation that incorporates the response.

\textbf{MP7. Audit emission.} The point at which the evidence record for the action is sealed. Mediation at MP7 evaluates the \textit{evidence record itself} --- its completeness, its cryptographic binding, its consistency with the decision projection. MP7 is the architecture's commitment point for accountability: once the evidence record is sealed and emitted to the audit substrate (Section 7), the action has been recorded as having occurred under specified authority and constraint.

The seven mediation points are not equal in cost or in defensive value. MP1 is the most powerful point --- earliest, broadest visibility, lowest realization cost --- but is also the point at which most current agent runtimes provide the least explicit hook. MP5 is the most commonly mediated point in existing systems, because it coincides with the tool-call API boundary, but it is also the point at which mediation cost is highest (the action is about to fire) and mediation richness is lowest (the system has less information about what the agent intended to do \textit{with} the action). A stop-anywhere mediator does not privilege any single point; it is present at all seven, and applies policy at the points where the policy's predicates can be evaluated.

\subsection{Mediation vs. Enforcement}

We distinguish two related but distinct properties of the system at any given mediation point.

\textbf{Mediation} is \textit{presence}: the system observes what is happening, evaluates relevant policy predicates, and produces a decision-projection update. Mediation is non-blocking in the common case --- the engine's evaluation is fast enough not to gate the agent's progress, and the engine's output is a decision projection that may simply confirm the proposed action.

\textbf{Enforcement} is \textit{action}: the system blocks, modifies, deflects, or rolls back. Enforcement is the subset of mediation outcomes that produce a non-identity decision projection. Of the six interruption primitives, allow (the degenerate case discussed in Section 5.7) is mediation without enforcement; the other five are mediation with enforcement of various kinds.

The distinction matters operationally. A system that is present only at points where it intends to enforce has no visibility into the agent's behavior at other points, and therefore cannot build the session state or accumulated context that subsequent policy decisions require. A stop-anywhere mediator must be present everywhere, even if it enforces nowhere --- because the audit substrate (Section 7) depends on the engine having observed the full trace of the agent's execution, not merely the points where it intervened.

We state this as a claim.

\begin{claim}[Mediation Subsumes Enforcement]
In a stop-anywhere mediator, the cost of mediation is amortized across the agent's session: every observed mediation point contributes to the session state that subsequent decisions consume. A system that mediates only at enforcement points carries the same per-decision cost but loses the cross-decision context that enables stateful policy evaluation.
\end{claim}

Claim 2 is the architectural justification for the cost of stop-anywhere mediation: presence at every point is not a tax, it is the substrate on which stateful policy stands.

\subsection{The Six Interruption Primitives}

The reasoning plane's output at each mediation point is drawn from a vocabulary of six structured primitives. We define each primitive formally, specify its input and output state, and indicate the runtime support it requires.

\subsubsection{Pause}

\begin{quote}
\textbf{Primitive 1 (Pause).} \textit{A synchronous interruption that suspends the agent's execution at the current mediation point until a specified condition resolves.}
\end{quote}

\textbf{Input.} Current execution state $\sigma$; resolution predicate $\phi$; optional timeout $\tau$.

\textbf{Output.} Paused state $\sigma_{\mathrm{paused}}$ with resumption token; on $\phi$ becoming true (or $\tau$ elapsing), execution resumes from $\sigma$ with $\phi$'s witness available to subsequent mediation.

\textbf{Required runtime support.} The agent runtime must expose a suspension primitive that captures execution state at the mediation point and admits resumption. In planner-executor runtimes, suspension at MP1, MP3, MP4, or MP6 is straightforward; suspension at MP2 or MP5 may require additional coordination with the data or network planes.

\textbf{Example use.} At MP5, the engine detects that a high-impact tool call is about to fire under a composite principal whose attenuation chain is borderline. Pause suspends the call until a parallel re-evaluation against the latest composite principal state completes.

\subsubsection{Escalate}

\begin{quote}
\textbf{Primitive 2 (Escalate).} \textit{An asynchronous interruption that routes the decision to a higher-authority adjudicator --- a human, a more-authorized agent, or an external workflow --- and resumes execution only upon that adjudicator's response.}
\end{quote}

\textbf{Input.} Current execution state $\sigma$; escalation context $\kappa$; target authority $\alpha$.

\textbf{Output.} Pending-approval state $\sigma_{\mathrm{pending}}$ registered with $\alpha$; the agent's execution is suspended pending $\alpha$'s response. Response is one of the six primitives applied to $\sigma$.

\textbf{Required runtime support.} The architecture must include a pending-approval substrate that registers the escalation, routes a notification to $\alpha$, and admits $\alpha$'s response as input to a continuation of the agent's execution. The substrate is conceptually a workflow engine \cite{Temporal,Cadence} with policy-aware routing.

\textbf{Example use.} At MP1, the engine detects that the agent's plan exceeds a risk threshold defined for the originating principal. Escalate routes the plan summary to a human approver; the agent's execution waits for an explicit accept or reject.

\subsubsection{Narrow}

\begin{quote}
\textbf{Primitive 3 (Narrow).} \textit{A synchronous interruption that allows the agent to proceed at the current mediation point but with an attenuated capability set, reflecting a strict subset of the composite principal's pre-mediation capabilities.}
\end{quote}

\textbf{Input.} Current execution state $\sigma$; attenuation directive $\delta$ (a capability-set subtraction).

\textbf{Output.} Continued execution at $\sigma'$ where the composite principal's effective capability set has been updated to the attenuated set; subsequent mediation points evaluate against the attenuated set.

\textbf{Required runtime support.} The composite principal must be a runtime-mutable object (Section 6) whose effective capability set can be updated mid-execution. The attenuation must propagate to all enforcement planes that consume the composite principal --- particularly the identity plane.

\textbf{Example use.} At MP3, the engine detects that the agent has selected a tool whose default permission scope is broader than necessary for the current plan step. Narrow attenuates the composite principal's effective capabilities for the duration of this tool invocation, so the tool's actual access is bounded to the policy-determined subset.

\subsubsection{Modify}

\begin{quote}
\textbf{Primitive 4 (Modify).} \textit{A synchronous interruption that rewrites the agent's proposed action --- the plan, the arguments, the retrieval query, or the response --- before it commits, returning to the agent the modified version as if it had been the original.}
\end{quote}

\textbf{Input.} Current execution state $\sigma$; proposed action $a$; rewriting function $r: a \mapsto a'$.

\textbf{Output.} Continued execution at $\sigma$ with $a'$ in place of $a$; the agent's downstream reasoning operates on $a'$.

\textbf{Required runtime support.} The runtime must permit the engine to substitute the agent's intended action with a modified action at the mediation point. In planner-executor runtimes this is straightforward at MP1 (plan rewriting) and MP4 (argument rewriting); at MP2 it requires the retrieval query rewriting hook; at MP6 it requires response rewriting prior to the agent's continuation.

\textbf{Example use.} At MP4, the engine detects that the agent has generated a tool call argument that references a sensitive raw identifier. Modify rewrites the argument to use a tokenized form, so the tool receives a reference rather than the raw value, and the agent's downstream reasoning operates on the tokenized form.

\subsubsection{Defer}

\begin{quote}
\textbf{Primitive 5 (Defer).} \textit{A synchronous interruption that allows the agent to proceed only when a specified downstream condition holds, returning execution to the present mediation point if the condition has not yet held by a specified deadline.}
\end{quote}

\textbf{Input.} Current execution state $\sigma$; deferral predicate $\phi$; deadline $d$.

\textbf{Output.} Deferred state $\sigma_{\mathrm{deferred}}$ with reactivation trigger on $\phi$; if $\phi$ holds before $d$, execution resumes from $\sigma$; if $d$ elapses, the engine re-evaluates and may issue a different primitive (frequently Pause or Escalate).

\textbf{Required runtime support.} A predicate registration substrate that watches for $\phi$ and triggers reactivation. This is conceptually similar to the escalation substrate but does not require routing to a human authority.

\textbf{Example use.} At MP5, the engine detects that the agent's action requires a downstream endpoint posture check that has not yet completed. Defer waits for the posture check; if posture is confirmed, the action proceeds.

\subsubsection{Rollback}

\begin{quote}
\textbf{Primitive 6 (Rollback).} \textit{A synchronous interruption that reverts the agent's execution to a prior checkpoint, executing compensation actions for any side-effecting operations that have already committed.}
\end{quote}

\textbf{Input.} Current execution state $\sigma$; target checkpoint $\sigma_0$; compensation transcript $C$ of side-effecting operations executed between $\sigma_0$ and $\sigma$.

\textbf{Output.} Restored state $\sigma_0$ after executing $C^{-1}$ (the inverse compensation actions, if available) for each side-effecting operation in $C$.

\textbf{Required runtime support.} The runtime must support checkpointing of agent execution state, and the enforcement planes (particularly the data plane) must expose compensation primitives for the side-effecting actions they realize. Rollback is the most demanding primitive on runtime support; its full realization depends on the broader workflow infrastructure \cite{SagaPattern,Temporal}.

\textbf{Example use.} At MP6, the engine detects that the tool's response indicates the action realized at MP5 has produced an unintended downstream effect. Rollback reverts to the pre-MP5 checkpoint and executes compensation actions for the realized effect.

We make two observations about the six primitives.

First, the primitives are \textit{composable}. Pause may resolve into Modify; Escalate may return a Narrow directive from the human approver; Defer may time out into Rollback. The architecture's decision projection (Section 4.2) admits sequential application of primitives across mediation points.

Second, the primitives are \textit{complete}. We argue (Section 5.7) that no policy decision a production-agent governance system requires falls outside the union of these six primitives. The argument is not a formal proof --- that would require a formal model of agentic actions and threats that the literature has not yet stabilized --- but a structural one: the seven mediation points and six primitives jointly admit every intervention pattern observed in production deployments and described in the recent agent-security literature \cite{SoKAgentic,PlanAndAct,TallamAuthProp}.

\subsection{State Preservation Requirements}

For the six primitives to be realizable at each of the seven mediation points, the runtime must preserve specified state at each point. We summarize the requirements in the form of a table that doubles as an implementation checklist for runtime authors.

\begin{table}[ht]
\centering
\small
\caption{State each of the seven mediation points must preserve, and the interruption primitives that state enables.}
\label{tab:state}
\begin{tabularx}{\textwidth}{lLL}
\toprule
Mediation point & State to capture & Required for primitives \\
\midrule
MP1 Plan formation & Plan AST; planner inputs; composite principal & All six \\
MP2 Context retrieval & Retrieval query; candidate set pre-ranking; data-plane directive & Modify, Defer, Narrow, Escalate, Pause \\
MP3 Tool selection & Tool identifier; pre-binding contract; composite principal & All six \\
MP4 Argument generation & Proposed argument set; argument provenance & Modify, Pause, Escalate, Narrow, Rollback \\
MP5 Action commit & Call payload; idempotency token; pre-commit checkpoint & All six (Rollback specifically requires the checkpoint) \\
MP6 Output return & Tool response; response provenance; agent's pre-incorporation state & All six \\
MP7 Audit emission & Decision projection; per-plane realization status; cryptographic binding & Modify (of the evidence record); Rollback only via post-emission compensation \\
\bottomrule
\end{tabularx}
\end{table}

The MP5 row carries the architecture's heaviest implementation burden: the pre-commit checkpoint is what makes Rollback realizable. We discuss this in Section 5.6.

\subsection{Compensation and Rollback}

The Rollback primitive is the design's most demanding requirement on the surrounding runtime. We treat it explicitly because the gap between the architecture's specification and current production runtimes' capability is largest here.

A rollback at MP5 or later requires that side-effecting operations executed against external systems be compensable: for each operation $o$ committed between checkpoint $\sigma_0$ and current state $\sigma$, the system must be able to execute an inverse operation $o^{-1}$ that restores the external state. The classical reference for this requirement is the \textit{saga pattern} \cite{GarciaMolinaSalem87,SagaPattern}, in which long-running transactions are decomposed into a sequence of local transactions with paired compensating transactions. Modern workflow engines \cite{Temporal,Cadence} --- and comparable systems such as AWS Step Functions and Camunda --- provide native support for saga execution.

For the reference architecture, we adopt the saga framing and require that the data-plane and network-plane directives in the decision projection include, where the directive is side-effecting, an explicit compensation directive. The reasoning plane treats compensation as part of policy, not as an afterthought: a policy clause that authorizes a side-effecting action under condition $\phi$ implicitly authorizes the compensation if $\phi$ later becomes false.

Three open problems remain in current implementations, which we discuss in Section 11.

First, \textit{not all side effects are compensable}. An email sent cannot be unsent; an external API call may produce effects that the calling system cannot reverse. The architecture handles this by classifying actions as compensable, partially compensable, or non-compensable at policy time, and by requiring explicit policy authorization for non-compensable actions.

Second, \textit{checkpoint cost is non-trivial}. Capturing the agent's pre-commit state at MP5 incurs storage and serialization cost that scales with the agent's context size. The architecture is most efficient when the runtime supports incremental checkpointing --- capturing only the delta from a prior checkpoint --- rather than full state snapshots.

Third, \textit{compensation under concurrency is hard}. If multiple agents are executing concurrently and their actions interleave at the data plane, the compensation for any single agent's rollback may need to coordinate with the other agents' actions. The classical solution is optimistic concurrency control with conflict detection \cite{Kung81}; we adopt this for the architecture without further specification.

\subsection{Why This Generalizes Allow/Deny}

We close the section by formalizing the claim that traditional Boolean policy evaluators are the degenerate case of a stop-anywhere mediator.

\begin{claim}[Subsumption of Boolean Evaluation]
Let $G$ be a stop-anywhere mediator restricted to (a) the single mediation point MP5 (action commit) and (b) the two-element interruption vocabulary $\{$Allow, Pause-indefinitely$\}$. Then $G$ is observationally equivalent to a traditional Boolean policy evaluator at the request-time boundary.
\end{claim}

The claim's force is twofold. First, it shows that the architecture is a strict generalization of existing systems: every policy expressible in OPA, Cedar, or a Zanzibar-derived system is expressible in the architecture, restricted to a single mediation point and two primitives. Second, it shows what existing systems \textit{cannot} express: policies that require mediation at points other than MP5, or that require any of the four interruption primitives (Narrow, Modify, Defer, Rollback) absent from the Boolean vocabulary.

We do not claim this generalization is free; richer expressivity carries implementation cost, runtime requirements, and operational complexity that Boolean evaluators avoid. The argument is structural: production-agent governance requires the richer vocabulary, and the architecture's cost is the cost of providing it.

We summarize the section's contribution as the following invariant.

\begin{invariant}[Mediation Coverage]
Under the reference architecture, every observable point in the agent's execution pipeline is a mediation point at which the reasoning plane is present and at which any of the six interruption primitives may be invoked. The architecture preserves agent utility by default (the engine's presence at a mediation point does not imply intervention) and preserves agent safety by construction (no execution point exists at which the engine is absent and at which a policy violation could therefore go undetected).
\end{invariant}

Invariant 2 is the design property's full statement. Sections 6 and 7 develop the substrate on which the invariant is maintained: the composite principal model that the reasoning plane evaluates against, and the audit substrate that records every mediation outcome and makes Invariant 2 observable.

\section{Composite Principals and Capability Attenuation}

Sections 4 and 5 established the architectural composition and the design property under which a single reasoning-plane decision is realized across the four infrastructure planes and at any of seven mediation points using any of six interruption primitives. The architecture's correctness, however, depends on a question we have so far left informally specified: \textit{against what does the reasoning plane evaluate policy?}

A traditional policy evaluator answers this question with an atomic principal: a user identifier, a service identifier, or a token-bearing client. Under composite delegation --- where an originating principal $u$ delegates a task to a planner agent $a_1$, which delegates a subtask to an executor agent $a_2$, which calls a tool that itself invokes a downstream agent $a_3$ --- the atomic-principal answer is structurally inadequate. Every existing answer to ``who is acting?'' returns one of the principals in this chain: $u$ in OAuth on-behalf-of flows that propagate the originating identity \cite{RFC8693}, $a_2$ or $a_3$ in service-mesh architectures that authenticate only the immediate caller, or some flattened composite in SAML assertion delegation. None of these answers preserves the information that the reasoning plane requires to evaluate policy correctly under composite delegation.

We develop in this section a formal model of the \textit{composite principal} as the canonical input to the reasoning plane's adjudication, define \textit{capability attenuation} as the architectural primitive that constrains how composite principals are constructed, and specify the adjudication algorithm by which the reasoning plane evaluates a proposed action against a composite principal. The model extends classical object-capability theory \cite{Miller06,Birgisson14} to the multi-agent runtime regime and builds directly on prior work in authorization propagation for multi-agent systems \cite{TallamAuthProp}.

\subsection{The Principal as a Chain, Not a Record}

Under composite delegation, the principal asserting authority for a proposed action is not a single identity but an ordered sequence of identities through which authority has flowed. We treat this sequence as a first-class runtime object.

\begin{definition}[Composite Principal]
A composite principal is an ordered sequence $\Pi = \langle \pi_0, \pi_1, \ldots, \pi_n \rangle$ where each $\pi_i$ is a tuple $(p_i, K_i, t_i, \tau_i)$ consisting of (a) a principal identifier $p_i$, (b) a capability set $K_i$ that $p_i$ holds at the time of delegation to $\pi_{i+1}$, (c) a delegation timestamp $t_i$, and (d) a time-to-live $\tau_i$ specifying the duration over which the delegation is valid. The sequence is ordered such that $\pi_0$ is the originating principal and $\pi_n$ is the acting principal at the point of adjudication.
\end{definition}

Three properties of the definition deserve emphasis.

First, the composite principal is \textit{structurally explicit}: every delegation step is a first-class element of the principal's representation, not a flattening of identities into a single ``effective'' principal. This is the property that lets the reasoning plane reason about delegation chains rather than only about the most recent caller.

Second, the composite principal \textit{carries its own history}: each $\pi_i$ records the capability set the corresponding principal \textit{held at the time of delegation}, not the principal's currently-held capabilities. This distinction matters because the principal's capabilities may change between $t_i$ and the adjudication time; the delegation should be evaluated against the capabilities the delegator could legitimately confer, not against capabilities the delegator may have since acquired or lost.

Third, the composite principal \textit{terminates at the acting principal}: $\pi_n$ is the identity that will physically commit the action at MP5 (Section 5.2.5). The acting principal's raw capabilities are not, by themselves, the input to adjudication; they are one term in a chain whose intersection determines the effective capability set against which the action is evaluated.

The composite principal is a runtime-mutable object. The Narrow primitive (Section 5.4.3) operates on it by appending an attenuation step that reduces the effective capability set for subsequent mediation points. The Modify primitive (Section 5.4.4) may rewrite the action without altering the composite principal. The Rollback primitive (Section 5.4.6) restores the composite principal to a prior state. We require that all mutations be observable in the audit substrate (Section 7).

\subsection{Capability Attenuation as Primitive}

The composite principal's structural explicitness creates a question: under what constraints may $\pi_{i+1}$'s capability set $K_{i+1}$ be constructed from $\pi_i$'s capability set $K_i$? The architecture's answer is a single primitive that we elevate from a policy convention to a structural property of the principal model.

\begin{definition}[Capability Attenuation]
Capability attenuation is the property that, for every delegation step from $\pi_i$ to $\pi_{i+1}$, the delegated capability set $K_{i+1}$ is a subset of the parent capability set $K_i$:*

\textit{\[K_{i+1} \subseteq K_i \quad \forall i \in \{0, 1, \ldots, n-1\}\]}

*The subset relation is enforced at delegation time, not at evaluation time, and is non-bypassable by composition: there exists no policy clause whose effect is to expand the child's capability set beyond the parent's.
\end{definition}

The definition has the deliberately strong property of being a structural constraint on the principal model rather than a policy expression. A policy clause that asserts a capability $k \in K_{i+1}$ but $k \notin K_i$ is not merely denied; it is rejected as malformed. This rejection is what makes attenuation non-bypassable: an adversary cannot construct a policy whose composition with another policy yields capability expansion, because no individual policy can express capability expansion at all.

We make three observations on this design choice.

First, attenuation-as-primitive is the architectural counterpart of the macaroon authorization model \cite{Birgisson14}, in which contextual caveats may only restrict, never expand, the bearer's authority. The reference architecture extends the macaroon insight from credential design to runtime principal evaluation: the composite principal is, in effect, a chain of macaroons, each of which strictly restricts the previous.

Second, attenuation-as-primitive forecloses the classical Confused Deputy problem \cite{Hardy88}. In the Confused Deputy, a process with elevated authority is induced by a lesser-authorized caller to perform an action on the caller's behalf, with the elevated authority's privileges. Under attenuation, the elevated authority cannot exceed the intersection of the chain's capability sets; the deputy's privileges are bounded by the caller's, by construction.

Third, attenuation-as-primitive is \textit{the} structural property of the composite principal that makes the model formally tractable. Without it, the composite principal's effective capability set is policy-dependent and cannot be precomputed; with it, the effective capability set is well-defined as the intersection of the chain's capability sets, and the adjudication algorithm (Section 6.5) reduces to evaluating policy against a single derived set.

We formalize the effective capability set explicitly.

\begin{definition}[Effective Capability Set]
Given a composite principal $\Pi = \langle \pi_0, \pi_1, \ldots, \pi_n \rangle$, the effective capability set $K^*(\Pi)$ at the time of adjudication is:*

\textit{\[K^*(\Pi) = \bigcap_{i=0}^{n} K_i\]}

*where each $K_i$ is the capability set recorded at delegation step $i$, restricted to the capabilities whose TTL $\tau_i$ has not yet elapsed.
\end{definition}

The intersection is monotonically non-expanding under further delegation: $K^*(\langle \Pi, \pi_{n+1} \rangle) \subseteq K^*(\Pi)$. This monotonicity is what bounds the blast radius of any compromised delegation: an adversary who compromises $\pi_k$ can act with at most $K^*(\langle \pi_0, \ldots, \pi_k \rangle)$, which is by construction a subset of every prior $K_i$ in the chain.

We adopt a partial-order view of capability sets under delegation, which we summarize formally.

\begin{claim}[Capability-Set Lattice]
Capability sets under attenuation form a monotone lattice under the subset partial order. Every delegation step is a downward move in the lattice; no policy-expressible operation produces an upward move. The lattice's top element is the union of all capabilities the originating principal $\pi_0$ holds; the lattice's bottom element is the empty set.
\end{claim}

The lattice structure is what gives the formal-methods community a hook into the architecture. Properties of the form ``no execution trace reaches an effective capability set $K^*$ that contains capability $k$ unless $k$ is held by every $\pi_i$ in the chain'' are decidable on the lattice and can be checked statically against policy specifications.

\subsection{Time-to-Live as First-Class}

Every capability in every $K_i$ carries an explicit time-to-live (TTL). This is not a constraint expressed in policy; it is a structural property of the capability itself.

\begin{definition}[Capability TTL]
Each capability $k \in K_i$ is a tuple $k = (\rho, \omega, \gamma, \tau)$ where $\rho$ is the resource, $\omega$ is the operation, $\gamma$ is the constraint set (a vector of predicates the action must satisfy), and $\tau$ is the time-to-live after which the capability is removed from $K_i$.
\end{definition}

Three structural properties of TTL deserve explicit statement.

First, TTL is \textit{non-extendable under delegation}. The child capability's TTL $\tau_{i+1}$ for a given capability is constrained to be no larger than the parent's $\tau_i$:

\[\tau_{i+1} \leq \tau_i\]

A delegation step may narrow the TTL --- the child may expire its inherited capability earlier than the parent --- but cannot extend it. This is the temporal analog of capability attenuation: just as the child's capability set is a subset of the parent's, the child's TTL on each capability is bounded above by the parent's.

Second, expired capabilities are \textit{removed from $K_i$, not merely flagged}. The reasoning plane evaluates only against unexpired capabilities; an expired capability is structurally absent from the effective capability set $K^*$. This makes TTL-based correctness arguments tractable: an expired capability cannot accidentally satisfy a policy clause because it is not in the set against which the clause is evaluated.

Third, TTL bounds the \textit{adversarial blast radius} of any compromised delegation. If $\pi_k$ is compromised at time $t_c$, the adversary's effective capability set is $K^*(\langle \pi_0, \ldots, \pi_k \rangle)$ restricted to capabilities whose TTL extends beyond $t_c$. Capabilities with $\tau_i$ ending before $t_c$ are unavailable to the adversary by construction. This is the architectural justification for short, structurally-bounded TTLs: they reduce the maximum duration over which any compromise is exploitable.

The architecture forbids open-ended delegations. Every $\tau_i$ must be finite. Long-lived authority is expressed not as a single delegation with large $\tau$ but as a renewal pattern: a controller re-delegates the capability at intervals, with each renewal observable in the audit substrate. This pattern preserves the architecture's correctness arguments while accommodating long-running agent workflows.

\subsection{The Adjudication Algorithm}

Given the composite principal model and the attenuation primitive, the reasoning plane's adjudication of a proposed action reduces to a well-defined algorithm. We state the algorithm formally; its complexity is linear in the length of the composite principal chain.

\begin{adjalg}[Composite Principal Adjudication]
\textit{Input.} A composite principal $\Pi = \langle \pi_0, \pi_1, \ldots, \pi_n \rangle$; a proposed action $a$; a policy $\mathcal{P}$; session state $s$.

\textit{Output.} A decision projection $D(a)$ (Section 4.2) drawn from the six interruption primitives.

\textit{Procedure.}

1. \textit{TTL filtering.} For each $\pi_i$ in $\Pi$, remove from $K_i$ every capability whose TTL has elapsed. Let $K_i'$ be the result.

2. \textit{Effective capability computation.} Compute $K^*(\Pi) = \bigcap_{i=0}^{n} K_i'$.

3. \textit{Capability check.} Determine whether the proposed action $a$ requires a capability $k$ such that $k \in K^*(\Pi)$. If not, return Pause-indefinitely (the degenerate Deny projection).

4. \textit{Constraint evaluation.} For each capability $k$ required by $a$, evaluate the constraint set $\gamma$ against the action's arguments. If any constraint fails, the engine selects from {Modify, Narrow, Pause, Escalate} according to policy.

5. \textit{Session-state evaluation.} Apply $\mathcal{P}$'s session-state predicates against $s$. If any predicate triggers, the engine selects an appropriate interruption primitive.

6. \textit{Decision projection.} Construct $D(a)$ by selecting the interruption primitive and projecting per-plane directives $\langle P_{\mathrm{net}}, P_{\mathrm{id}}, P_{\mathrm{ep}}, P_{\mathrm{data}} \rangle$ that realize the primitive across the four infrastructure planes.

7. \textit{Evidence record.} Construct the evidence record $e$ binding the composite principal, the decision projection, and the policy clauses invoked.
\end{adjalg}

The algorithm has three correctness properties worth stating explicitly.

\begin{claim}[Adjudication Correctness]
Algorithm 1 satisfies the following properties:*

\textit{(a) No action is authorized except against a capability in $K^*(\Pi)$.}
\textit{(b) No expired capability contributes to authorization.}
*(c) The decision projection is reproducible: given the same $\Pi$, $a$, $\mathcal{P}$, and $s$, the algorithm produces the same $D(a)$.
\end{claim}

The three properties follow directly from the construction of Algorithm 1. Property (a) holds because step 3 tests the action's required capability against $K^*(\Pi)$ and returns the Deny projection when the test fails; no later step can reintroduce a capability absent from $K^*(\Pi)$, since steps 4--6 only further constrain. Property (b) holds because step 1 removes every expired capability from each $K_i$ before step 2 computes $K^*(\Pi) = \bigcap_i K_i'$, so an expired capability is structurally absent from the set against which authorization is tested. Property (c) holds because the procedure reads only its inputs $\langle \Pi, a, \mathcal{P}, s\rangle$ and contains no nondeterministic step: the same inputs traverse the same branches and produce the same $D(a)$.

Property (a) is thus the chain-bounded authority property: actions are limited to capabilities present in every link of the delegation chain. Property (b) is the TTL property: temporal bounds on capabilities are enforced at evaluation time. Property (c) is the determinism property, which is what makes the audit substrate's evidence records reconstructible (Section 7.2) and what makes the architecture's behavior verifiable.

Properties (a) and (b) jointly bound the maximum authority any composite principal can exercise at any moment. Together with the lattice structure (Claim 4), they make the architecture's correctness arguments amenable to model-checking and to formal verification of policy specifications, which we discuss further in Section 11.

\subsection{Comparison with Prior Delegation Models}

The composite principal model extends and improves on three prior delegation models. We compare each.

\subsubsection{OAuth On-Behalf-Of and Token Exchange}

OAuth 2.0 on-behalf-of flows \cite{RFC8693,OAuthOBO} permit a client to acquire a new token that asserts authority delegated from an upstream principal. The acquired token carries claims about the originating principal but does not natively represent the chain through which authority was delegated. In particular, attenuation is expressed only by the requested scope at token-exchange time and is not enforced structurally: a client may request and receive a token with scopes that exceed what an intermediary in the delegation chain held.

The composite principal model differs in three respects. First, the chain is explicit at evaluation time; it is not flattened into a single derived token. Second, attenuation is structural, not requested: $K_{i+1} \subseteq K_i$ is enforced at delegation time and cannot be violated even by an adversarial intermediary. Third, TTL is per-capability, not per-token: individual capabilities may expire while the token remains valid for others.

\subsubsection{SAML Assertion Delegation}

SAML 2.0 assertion delegation \cite{SAMLdelegation} uses the \texttt{<DelegationRestriction>} and \texttt{<Delegate>} elements to express that an assertion is issued on behalf of a chain of prior subjects. The chain is explicit, which is the SAML model's strength. However, SAML expresses delegation at the assertion level rather than the capability level: an assertion conveys identity and a coarse set of attributes, with no formal model of capability attenuation across the chain. Authorization is left to the application, which typically flattens the chain to one or another of its subjects.

The composite principal model adopts SAML's explicit-chain insight and combines it with macaroon-style attenuation at the capability level, producing a representation that is both structurally explicit \textit{and} formally attenuated.

\subsubsection{Zanzibar and Relationship-Based Authorization}

Zanzibar \cite{Pang19} expresses authorization as a graph of relationships among users, groups, and resources. Computed usersets and caveats permit policy expressions that conditionally derive new relationships. Zanzibar's model is exceptionally well-suited to \textit{static} relationship structures --- organizational hierarchies, sharing graphs, group memberships --- but does not natively model \textit{dynamic delegation chains} in which the chain itself is constructed at runtime and whose authority bounds are determined by the chain's structure, not by a precomputed graph.

The composite principal model is complementary to Zanzibar rather than competitive with it. Zanzibar is the appropriate substrate for the \textit{base} capability sets of identifiable principals; the composite principal model determines how those base capability sets compose under runtime delegation. A production implementation may use a Zanzibar-derived system \cite{SpiceDB,Permify} for base capability evaluation and the composite principal model for chain-level adjudication.

\subsubsection{Macaroons}

Macaroons \cite{Birgisson14} are bearer credentials with contextual caveats that may only restrict, never expand, the bearer's authority. The macaroon model is the closest prior art to the composite principal: it pioneered attenuation-as-primitive at the credential layer.

The composite principal model extends the macaroon insight in three ways. First, it operates at the runtime principal layer rather than the credential layer, integrating with policy evaluation at adjudication time rather than at credential presentation. Second, it preserves the explicit chain, so policy may reason about who delegated to whom under what bounds, not merely about the final attenuated set. Third, it integrates TTL as a per-capability property, where macaroons typically rely on a \texttt{time\_before} caveat applied to the entire credential.

In a slogan: macaroons are the architecture's intellectual ancestor at the credential layer, and the composite principal is the architecture's extension of that ancestry to the runtime layer.

\subsection{Cryptographic Binding and Tamper Evidence}

The composite principal model is operationally meaningful only if the chain is tamper-evident: an adversary who acquires access to the composite principal's representation must be unable to extend the chain, remove an attenuation step, or alter a capability set in a way that escapes detection at adjudication time.

We require that the composite principal be cryptographically bound. Each delegation step $\pi_i$ is signed by the delegating principal $p_{i-1}$ (or, for $\pi_0$, by the issuing authority). The signature covers the principal identifier, the capability set, the timestamp, and the TTL. The composite principal as a whole is a chain of signatures; an adversary who attempts to alter $\pi_k$ for $k < n$ must forge $p_{k-1}$'s signature.

This construction is the macaroon scheme extended to multi-step delegation \cite{Birgisson14}; the implementation considerations (key management, signature scheme choice, signature size) follow the macaroon literature and are not novel to this architecture. We address the formal-verification implications and the key management open problems in Section 11.

The cryptographic binding is what makes the audit substrate's evidence records meaningful. The evidence record (Section 7) includes the full composite principal as part of its sealed content; without cryptographic binding, the audit substrate would record an adjudication against a principal whose authenticity cannot be verified post hoc.

We summarize the section's contribution as the following invariant.

\begin{invariant}[Bounded Composite Authority]
Under the reference architecture, every action realized through the architecture is bounded by the effective capability set $K^*(\Pi)$ of its composite principal $\Pi$. The architecture admits no execution trace in which an action is realized against a capability outside $K^*(\Pi)$, and no execution trace in which the composite principal is altered without producing a verifiable signature failure at adjudication time.
\end{invariant}

Invariant 3 is the formal statement of the architecture's authority-correctness property. It is what makes the architecture defensible to a security reviewer who asks: ``what bounds the authority of any agent in your system?'' The answer is: the intersection of the capability sets along its delegation chain, restricted to unexpired capabilities, cryptographically bound to the chain itself.

\section{Audit as a Structured Evidence Substrate}

The architecture developed in Sections 4 through 6 makes a sequence of decisions --- one per agent action, each adjudicated at the reasoning plane against a composite principal and realized across the four infrastructure planes. For these decisions to be governable in the sense regulated enterprises require, the system must produce more than a record that an action occurred. It must produce \textit{evidence}: a structured, reconstructible, tamper-evident account of who acted, under what delegated authority, against what data, under what policy, with what outcome, and across which enforcement planes the decision was realized.

We argue in this section that the audit function in a production-agent governance system is not logging but \textit{evidence production}, and that this distinction has architectural consequences. We define the structure of the evidence record, specify the property of reconstructability under partial information --- drawing on prior work in benchmarking authorization-limited evidence \cite{TallamPartialEvid} --- and show how the evidence substrate composes across the five planes without itself becoming a point of vendor lock-in.

\subsection{Audit versus Logging}

We begin by distinguishing two functions that production systems conflate.

\textit{Logging} is the unstructured or semi-structured emission of events for human consumption and post-hoc debugging. A log entry is optimized for a developer reading it during an incident: it is free-text or loosely-structured, it is emitted at the discretion of the code path that produces it, and it carries no guarantee of completeness, consistency, or tamper-evidence. The dominant operational practice in production AI systems treats audit as a specialization of logging --- a log stream tagged for compliance.

\textit{Audit}, in the sense this architecture requires, is the structured emission of evidence for adjudication. An audit record is optimized not for a developer reading it during an incident but for an auditor, a regulator, or an incident responder reconstructing what the system did and under whose authority, potentially long after the fact and potentially with only partial access to the system's state. The distinction has four architectural consequences.

First, audit records are \textit{structured by contract}, not by convention. Every evidence record carries the same schema (Section 7.2), so that records are queryable, comparable, and machine-analyzable across the entire system.

Second, audit records are \textit{complete by construction}. In the federated-enforcement model criticized in Section 4, each plane emits its own log stream, and no single record captures the full decision. Under the reference architecture, the reasoning plane produces one evidence record per decision, and every enforcement plane annotates that record with its realization status. Completeness is a property of the architecture, not of developer diligence.

Third, audit records are \textit{tamper-evident}. The evidence record is cryptographically bound (Section 7.4), so that an auditor can verify that a record has not been altered since emission. Logs offer no such guarantee.

Fourth, audit records are \textit{reconstructible under partial information}. An auditor with access to only a subset of the system's evidence should be able to bound what the system did, even if the full record set is unavailable. This property --- which we develop in Section 7.3 --- is the audit substrate's hardest requirement, and the one most distinctive to the agentic regime.

We state the distinction as a principle.

\begin{principle}[Audit as Evidence]
A production-agent governance system must treat audit as evidence production, not as logging. Evidence is structured by contract, complete by construction, tamper-evident by cryptographic binding, and reconstructible under partial information. A system that emits logs and tags them for compliance does not satisfy this principle.
\end{principle}

\subsection{The Evidence Record}

The unit of audit in the architecture is the evidence record, produced once per decision by the reasoning plane and composed with per-plane realization status at the fan-in step (Section 4.4).

\begin{definition}[Evidence Record]
An evidence record for an action $a$ is a tuple*

\textit{\[e(a) = \langle \Pi, D(a), \mathcal{C}, R, \chi, \mu \rangle\]}

*where $\Pi$ is the composite principal at adjudication time (Section 6), $D(a)$ is the decision projection (Section 4.2), $\mathcal{C}$ is the set of policy clauses invoked in the decision, $R = \langle R_{\mathrm{net}}, R_{\mathrm{id}}, R_{\mathrm{ep}}, R_{\mathrm{data}} \rangle$ is the per-plane realization status, $\chi$ is the set of correlation identifiers binding this record to the broader call chain, and $\mu$ is the cryptographic binding (Section 7.4).
\end{definition}

We comment on each component.

\textbf{The composite principal $\Pi$.} The evidence record carries the \textit{full} composite principal --- the complete delegation chain, not merely the acting principal. This is what allows an auditor to answer the question ``under whose authority was this action taken?'' with the full chain: the originating principal, every intermediary, and the attenuation steps that bounded the effective capability set. An audit record that carries only the acting principal cannot answer this question, and is therefore insufficient for the agentic regime.

\textbf{The decision projection $D(a)$.} The evidence record carries the reasoning plane's full decision, including which of the six interruption primitives was selected (Section 5.4) and the per-plane directives. This is what allows an auditor to answer ``what did the system decide to do?'' --- including the cases where the system intervened (modified an argument, narrowed a capability set, escalated to a human), which an outcome-only record would not capture.

\textbf{The policy clauses $\mathcal{C}$.} The evidence record carries the specific policy clauses that produced the decision. This is what allows an auditor to answer ``why was this decision made?'' --- and what allows a policy author to trace a contested decision back to the clause responsible. The determinism property (Claim 5c, Section 6.4) ensures that re-evaluating the recorded $\Pi$, $a$, and $\mathcal{C}$ reproduces $D(a)$, so that decisions are not merely recorded but \textit{reproducible}.

\textbf{The realization status $R$.} The evidence record carries each enforcement plane's report of whether it realized its directive. This is what makes Invariant 1 (Composed Authority, Section 4.4) observable: an auditor can verify that every plane that should have realized a directive did so, and that the realizations are jointly consistent with the reasoning plane's decision.

\textbf{The correlation identifiers $\chi$.} The evidence record carries identifiers that bind it to the broader call chain --- the originating request, the parent agent's evidence record, the downstream tool's evidence record. This is what allows reconstruction of multi-step and multi-agent workflows from their constituent records.

\textbf{The cryptographic binding $\mu$.} The evidence record is sealed with a cryptographic binding that makes alteration detectable. We develop this in Section 7.4.

The evidence record is emitted at MP7 (audit emission, Section 5.2.7), the final mediation point. The Modify primitive may be applied to the evidence record itself at MP7 --- for instance, to redact a field that policy forbids from being recorded in plaintext --- but no other primitive alters a sealed record. Once emitted, the record is immutable; corrections take the form of subsequent records that reference the original, never of in-place modification.

\subsection{Reconstructability Under Partial Information}

The audit substrate's most distinctive requirement is that evidence be reconstructible even when only a subset of records is available. This requirement arises directly from the agentic regime: a multi-agent workflow may distribute evidence records across multiple systems, organizational boundaries, or retention domains, and an auditor reconstructing the workflow after the fact may have access to only some of them.

We draw on prior work that formalizes and benchmarks this property \cite{TallamPartialEvid}. The central construct is the \textit{partial evidence reconstruction}: given a subset $E' \subseteq E$ of the full evidence set $E$ for a workflow, an auditor should be able to compute a bound on what the workflow did, with the bound tightening monotonically as $E'$ grows toward $E$.

\begin{definition}[Reconstructability]
An audit substrate is reconstructible iff, for every workflow with full evidence set $E$ and every subset $E' \subseteq E$, there exists a reconstruction function $\mathrm{Rec}(E')$ that produces a sound over-approximation of the workflow's behavior: every action the workflow actually took is consistent with $\mathrm{Rec}(E')$, and $\mathrm{Rec}(E') \supseteq \mathrm{Rec}(E'')$ whenever $E' \subseteq E''$ (the reconstruction tightens monotonically as more evidence becomes available).
\end{definition}

The soundness requirement --- that the reconstruction is an over-approximation, never an under-approximation --- is what makes partial evidence useful for accountability. An auditor with partial evidence may not know exactly what the workflow did, but can be certain that the workflow did nothing outside the reconstructed bound. This is the property that allows an incident responder to bound a blast radius from incomplete evidence: even if some records are unavailable, the available records constrain what could have happened.

Reconstructability has three architectural consequences for evidence record design.

First, each evidence record must be \textit{self-describing}: it must carry enough context (the composite principal, the decision projection, the correlation identifiers) to be interpreted in isolation, without requiring access to other records. A record that can only be understood in the context of records that may be unavailable does not support reconstruction.

Second, the correlation identifiers $\chi$ must form a \textit{navigable structure}: from any record, an auditor must be able to identify which other records are its parents, children, and siblings in the workflow, even if those records are not themselves available. This lets the auditor know what evidence is missing, which is itself information that bounds the reconstruction.

Third, the evidence substrate must distinguish \textit{absence of evidence} from \textit{evidence of absence}. A reconstruction must be able to represent ``this record is missing'' distinctly from ``no action occurred here,'' because conflating the two produces unsound reconstructions. This is the property that the \textit{Partial Evidence Bench} benchmark \cite{TallamPartialEvid} is designed to measure: how well a system's evidence supports sound reconstruction under varying degrees of evidence availability.

These three consequences are design requirements; that the architecture's evidence records (Definition 7) actually satisfy Definition 8 is a claim we substantiate empirically rather than by proof. The reference implementation (Section 10.8) supplies a concrete reconstruction function over the architecture's evidence records and confirms, across a thousand synthetic workflows, that the reconstruction is sound on every trial --- it never under-approximates the workflow's true action set --- and monotonic on every trial --- it never shrinks as more records become available. The architecture is thus reconstructible in the sense of Definition 8 not merely by design but as measured behavior.

\subsection{Cryptographic Binding and Tamper Evidence}

For the evidence record to serve as evidence --- in the accountability sense, potentially in an adversarial setting --- it must be tamper-evident: any alteration to a sealed record must be detectable, and the integrity of the record set as a whole must be verifiable.

We require two cryptographic properties.

\textbf{Per-record integrity.} Each evidence record is signed at emission, with the signature covering the entire record tuple $\langle \Pi, D(a), \mathcal{C}, R, \chi \rangle$. The signing authority is the reasoning plane instance that produced the decision. An auditor verifies that a record has not been altered since emission by checking the signature against the reasoning plane's public key.

\textbf{Set integrity.} The evidence records for a workflow are chained into a tamper-evident structure --- a hash chain or Merkle structure --- so that the removal or reordering of records is detectable. This is what prevents an adversary who has compromised the audit store from silently deleting records: removing a record breaks the chain, and the break is detectable by any party holding a later record or a periodic attestation of the chain's head.

The cryptographic constructions here are standard \cite{Merkle80,Haber91}; we do not claim novelty in the cryptographic substrate. What is specific to the architecture is the \textit{content} that is bound: the composite principal (itself cryptographically bound per Section 6.6), the decision projection, the policy clauses, and the per-plane realization status. The evidence record's tamper-evidence is meaningful precisely because the bound content is the full account of the decision, not a summary.

We note one consequence for retention. Because the evidence record set is a tamper-evident chain, retention policies must preserve enough of the chain to maintain verifiability. A retention policy that deletes old records must either preserve the chain's integrity proofs (e.g., by retaining Merkle roots and inclusion proofs for deleted records) or explicitly accept a verifiability boundary before which set integrity can no longer be checked. We treat the design of retention-compatible tamper-evidence as an implementation concern, addressed by the broader literature on long-term secure logging \cite{Schneier99,Holt06}.

\subsection{Composability Across Planes}

The evidence record is produced by the reasoning plane but completed by the enforcement planes. We specify how the record composes across the five planes without requiring the planes to share infrastructure.

At the fan-out step (Section 4.4), the reasoning plane dispatches per-plane directives and creates a provisional evidence record carrying $\Pi$, $D(a)$, $\mathcal{C}$, and $\chi$, with the realization status $R$ unpopulated. Each enforcement plane, upon realizing or failing to realize its directive, emits a \textit{realization sub-record} --- a small, signed attestation of its outcome --- referencing the provisional evidence record via the correlation identifiers $\chi$. At the fan-in step, the audit substrate composes the provisional record with the realization sub-records to produce the completed, sealed evidence record.

This composition has a useful property: each plane's sub-record is independently meaningful and independently verifiable. An auditor with access only to the data plane's realization sub-records can verify what the data plane realized, even without access to the reasoning plane's records --- and can, via the correlation identifiers, determine what reasoning-plane records would be needed to complete the picture. This is the plane-level instantiation of reconstructability (Section 7.3): the evidence substrate degrades gracefully under partial access, with each plane contributing independently verifiable evidence.

The architectural significance is that the audit substrate is an \textit{emission contract}, not a storage system. The reasoning plane and the four enforcement planes each emit evidence according to the contract; where that evidence is stored, queried, retained, and analyzed is outside the architecture's scope.

\subsection{Integration with Existing Infrastructure}

A production deployment of the architecture does not require a bespoke audit store. The evidence record's value comes from its structure and its cryptographic binding, not from any particular storage or analysis platform. We specify the integration boundary explicitly to forestall the misreading that the audit substrate is a monolithic component the enterprise must adopt wholesale.

The evidence record is emitted as a structured event conforming to the contract of Definition 7. Existing security information and event management systems \cite{Splunk,Sentinel,Chronicle}, observability pipelines \cite{OpenTelemetry,Datadog,Honeycomb}, and compliance-evidence platforms can consume these events as sinks. The evidence record's correlation identifiers $\chi$ are designed to align with the trace-context conventions of distributed-tracing systems \cite{W3CTraceContext}, so that an evidence record can be correlated with the distributed trace of the agent's execution in the enterprise's existing observability stack.

This integration boundary serves two purposes. First, it avoids vendor lock-in on the audit substrate: the enterprise retains its existing SIEM, observability, and compliance tooling, and the architecture contributes structured, signed evidence into those systems rather than replacing them. Second, it preserves the enterprise's existing retention, access-control, and analysis policies for audit data, which are frequently subject to their own regulatory constraints that the architecture should not override.

We state the integration principle.

\begin{principle}[Audit as Emission Contract]
The audit substrate is an emission contract, not a storage or analysis system. The reference architecture specifies the structure, completeness, tamper-evidence, and reconstructability of evidence records; it does not specify where those records are stored, how they are queried, or how long they are retained. These remain the enterprise's existing concerns, into which the architecture emits structured evidence.
\end{principle}

We summarize the section's contribution as the following invariant.

\begin{invariant}[Evidence Sufficiency]
Under the reference architecture, every decision produces an evidence record that is sufficient to answer, for that decision: who acted (the composite principal), under what authority (the effective capability set and its delegation chain), against what (the resources and data touched), under what policy (the clauses invoked), with what outcome (the interruption primitive and per-plane realization), and bound such that alteration is detectable. Where evidence is partially available, the available records bound the system's behavior soundly.
\end{invariant}

Invariant 4 is the audit substrate's full statement. Together with Invariant 1 (Composed Authority, Section 4.4), Invariant 2 (Mediation Coverage, Section 5.7), and Invariant 3 (Bounded Composite Authority, Section 6.6), it completes the architecture's four-invariant correctness foundation. Section 8 composes the primitives developed in Sections 4 through 7 into a single reference architecture and traces a complete agent action through it.

\section{Reference Architecture: Composing the Primitives}

Sections 4 through 7 developed four primitives in isolation: the five-plane structural composition, stop-anywhere mediation across seven points with six interruption primitives, the composite principal model with capability attenuation, and the audit substrate that produces reconstructible evidence. We now compose these primitives into a single reference architecture and demonstrate the composition by tracing one complete agent action through it, end to end.

The purpose of this section is twofold. First, it makes the architecture concrete: the four primitives, each defined abstractly, are shown operating together on a single decision, so that the reader can see how the decision projection flows from adjudication to enforcement to evidence. Second, it surfaces the architecture's operational properties --- latency, policy-language requirements, and operability --- that are not visible when the primitives are considered in isolation but become load-bearing when they are composed into a deployable system.

\subsection{The Composed Architecture}

We summarize the architecture as a layered system with a single adjudication locus and four enforcement substrates, bound together by the decision projection (the forward contract) and the evidence record (the backward contract).

The components are:

\begin{itemize}
\item \textbf{The agent runtime} --- the planner, executor, and tool-invocation machinery that produces the agent's behavior. The runtime exposes the seven mediation points (Section 5.2) as hooks at which the reasoning plane is present.
\end{itemize}

\begin{itemize}
\item \textbf{The reasoning plane} --- the stateful adjudicator (Section 4.2) that evaluates each proposed action against the composite principal (Section 6), the agent's plan, the session state, and the policy, producing a decision projection (Section 4.2) drawn from the six interruption primitives (Section 5.4).
\end{itemize}

\begin{itemize}
\item \textbf{The four enforcement planes} --- network, identity, endpoint, and data (Section 4.3), each of which realizes its directive from the decision projection using its native enforcement primitives and emits a realization sub-record.
\end{itemize}

\begin{itemize}
\item \textbf{The composite principal store} --- the runtime-mutable representation of the delegation chain (Section 6.1), cryptographically bound (Section 6.6), against which the reasoning plane adjudicates.
\end{itemize}

\begin{itemize}
\item \textbf{The audit substrate} --- the emission contract (Section 7.6) that composes the reasoning plane's provisional evidence record with the enforcement planes' realization sub-records into a sealed, tamper-evident evidence record (Section 7.2).
\end{itemize}

\begin{itemize}
\item \textbf{The policy store} --- the versioned, declarative policy specification that the reasoning plane evaluates. We treat the policy language abstractly here and discuss its requirements in Section 8.4.
\end{itemize}

The forward path through the system, for a single action, is: the agent runtime presents a proposed action at a mediation point; the reasoning plane adjudicates it against the composite principal and policy, producing a decision projection; the decision projection fans out to the four enforcement planes; each plane realizes its directive; the realizations fan in to the audit substrate. The backward path is the evidence: each plane's realization sub-record, composed with the reasoning plane's decision, produces the sealed evidence record.

We make explicit the architecture's central structural commitment, which the composition now makes visible: \textit{adjudication happens once, at the reasoning plane, and is realized many times, across the enforcement planes, with a single composed evidence record binding the decision to its realizations.} This is the property that distinguishes the architecture from federated enforcement, in which adjudication happens independently at each plane and no single evidence record binds the decision together.

\subsection{End-to-End Trace of a Single Action}

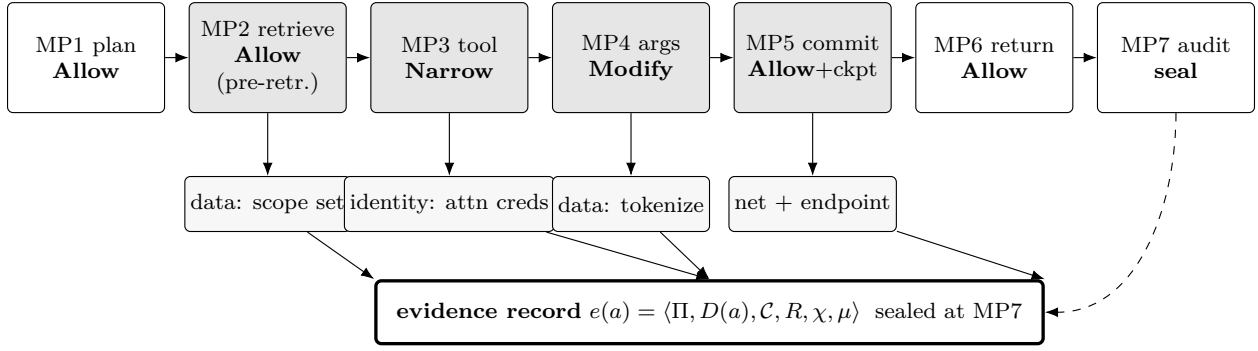
\begin{figure}[t]
\centering
\resizebox{\textwidth}{!}{%
\begin{tikzpicture}[
  >=Latex, font=\scriptsize,
  tstep/.style = {draw, rounded corners=2pt, align=center, inner sep=3pt,
                 minimum width=2.0cm, minimum height=14mm},
  act/.style  = {fill=black!10},
  plane/.style= {draw, rounded corners=2pt, fill=black!3, align=center,
                 inner sep=2pt, minimum width=2.0cm, minimum height=7mm},
  arr/.style  = {-Latex}
]
\node[tstep]            (s1) {MP1 plan\\\textbf{Allow}};
\node[tstep, act, right=3mm of s1] (s2) {MP2 retrieve\\\textbf{Allow}\\(pre-retr.)};
\node[tstep, act, right=3mm of s2] (s3) {MP3 tool\\\textbf{Narrow}};
\node[tstep, act, right=3mm of s3] (s4) {MP4 args\\\textbf{Modify}};
\node[tstep, act, right=3mm of s4] (s5) {MP5 commit\\\textbf{Allow}+ckpt};
\node[tstep, right=3mm of s5]      (s6) {MP6 return\\\textbf{Allow}};
\node[tstep, right=3mm of s6]      (s7) {MP7 audit\\\textbf{seal}};
\foreach \a/\b in {s1/s2,s2/s3,s3/s4,s4/s5,s5/s6,s6/s7}\draw[arr] (\a)--(\b);

\node[plane, below=8mm of s2] (pd) {data: scope set};
\node[plane, below=8mm of s3] (pi) {identity: attn creds};
\node[plane, below=8mm of s4] (pd2){data: tokenize};
\node[plane, below=8mm of s5] (pn) {net + endpoint};
\draw[arr] (s2)--(pd); \draw[arr] (s3)--(pi);
\draw[arr] (s4)--(pd2);\draw[arr] (s5)--(pn);

\node[draw, very thick, rounded corners=2pt, below=6mm of pd2, xshift=1.0cm,
      minimum width=8.5cm, minimum height=8mm, align=center]
      (ev) {\textbf{evidence record} $e(a)=\langle \Pi, D(a), \mathcal{C}, R, \chi, \mu\rangle$ \;sealed at MP7};
\draw[arr] (pd)--(ev.north west);
\draw[arr] (pi)--(ev.north);
\draw[arr] (pd2)--(ev.north);
\draw[arr] (pn)--(ev.north east);
\draw[arr, dashed] (s7) to[out=-90,in=0] (ev.east);
\end{tikzpicture}%
}
\caption{End-to-end trace of a single agent action through the architecture.}
\label{fig:trace}
\end{figure}

We trace one agent action through the architecture; Figure\textasciitilde{}\ref{fig:trace} depicts the trace end to end. The action is deliberately ordinary --- a production agent retrieving a document and incorporating it into a draft --- so that the trace illustrates the architecture's behavior on the common case, not only on adversarial edge cases.

\textbf{Scenario.} A production agent operates on behalf of a user $u$ who has delegated to it the task of drafting a summary of a quarterly report. The agent's plan requires retrieving the report from a document store, summarizing it, and writing the summary into a draft document. The composite principal at the start of the task is $\Pi = \langle \pi_0 \rangle$ where $\pi_0 = (u, K_u, t_0, \tau_0)$ and $K_u$ contains, among other capabilities, the capability to read documents in the finance workspace and the capability to write to the user's draft folder.

We trace the action of \textit{retrieving the quarterly report}.

\textbf{MP1 --- Plan formation.} The agent commits to a plan: retrieve report $\to$ summarize $\to$ write draft. The reasoning plane is present at MP1. It evaluates the plan as a whole against the composite principal $\Pi$ and the policy. The plan requires the read-finance-document and write-draft capabilities, both present in $K^*(\Pi) = K_u$. The plan also triggers a session-state predicate: the policy specifies that any plan that reads finance documents and writes to a folder must not, in the same session, write to an externally-shared folder. The current plan writes only to the user's private draft folder, so the predicate is satisfied. The reasoning plane's decision at MP1 is Allow (the degenerate primitive): the plan may proceed. An evidence record for the plan-formation decision is begun.

\textbf{MP2 --- Context retrieval.} The agent issues a retrieval for the quarterly report. The reasoning plane is present at MP2. It computes the data-plane directive $P_{\mathrm{data}}$: the retrieval candidate set is constrained to documents in the finance workspace that the composite principal is authorized to read. This is the pre-retrieval gate: the constraint is applied to the candidate set \textit{before} the documents reach the agent, so the agent cannot retrieve a document outside its effective capability set even if the retrieval query is broad. The decision is Allow, with the data-plane directive constraining the candidate set. The data plane realizes the directive by scoping the retrieval; it emits a realization sub-record indicating which candidate set was permitted.

\textbf{MP3 --- Tool selection.} The agent selects the document-read tool. The reasoning plane is present at MP3. The tool requires the read-document capability, which is in $K^*(\Pi)$. However, the policy specifies that the document-read tool, when invoked on a finance-workspace document, must operate under an attenuated capability set that excludes the write-external capability for the remainder of the session --- a separation-of-duty constraint preventing a read-then-exfiltrate chain. The reasoning plane's decision is Narrow (Section 5.4.3): the agent may proceed, but the composite principal is attenuated by appending $\pi_1 = (a, K_a, t_1, \tau_1)$ where $K_a = K_u \setminus \{\text{write-external}\}$. The identity plane realizes the Narrow directive by issuing short-lived credentials reflecting the attenuated set. An evidence record captures the attenuation.

\textbf{MP4 --- Argument generation.} The agent generates the arguments for the document-read tool: the document identifier and the requested fields. The reasoning plane is present at MP4. The policy specifies that finance documents may be read but that certain fields --- a set of personally-identifying fields --- must be tokenized before they reach the agent. The reasoning plane's decision is Modify (Section 5.4.4): the argument set is rewritten so that the tokenizable fields are requested in tokenized form. The agent's downstream reasoning will operate on tokenized values. An evidence record captures the modification.

\textbf{MP5 --- Action commit.} The document-read call is about to cross the system boundary. The reasoning plane is present at MP5. All prior decisions are consistent; the composite principal is attenuated as required; the arguments are modified as required. The reasoning plane creates a pre-commit checkpoint (so that Rollback is available if the response later reveals a problem) and issues Allow. The network plane realizes its directive (the call may reach the document store under the required cryptographic posture); the endpoint plane confirms the runtime posture; the data plane realizes the tokenization. The call fires. Realization sub-records are emitted by each plane.

\textbf{MP6 --- Output return.} The document store returns the report (with the personally-identifying fields tokenized). The reasoning plane is present at MP6. It evaluates the response: the content is within the authorized scope, the tokenization was applied, no anomalous provenance markers are present. The decision is Allow. The agent incorporates the response and proceeds to the summarization step (which would begin its own MP1).

\textbf{MP7 --- Audit emission.} The evidence record for the retrieval action is sealed. It carries: the full composite principal at adjudication time, including the attenuation step $\pi_1$; the decision projection across all mediation points (Allow at MP1, MP2, MP5, MP6; Narrow at MP3; Modify at MP4); the policy clauses invoked at each point; the per-plane realization status (data plane: candidate-set constraint realized and tokenization applied; identity plane: attenuated credentials issued; network plane: call realized under required posture; endpoint plane: posture confirmed); the correlation identifiers binding this record to the plan-formation record and to the forthcoming summarization record; and the cryptographic binding. The record is emitted to the audit substrate.

The trace illustrates four properties that are not visible when the primitives are considered in isolation. First, \textit{the common case exercises multiple interruption primitives}: even this ordinary retrieval invoked Narrow and Modify, not merely Allow/Deny. Second, \textit{the composite principal mutates mid-action}: the attenuation at MP3 changes the effective capability set for the rest of the session. Third, \textit{the four enforcement planes realize different parts of a single decision}: data, identity, network, and endpoint each realized a portion of the reasoning plane's projection. Fourth, \textit{the evidence record is the composition of all of this}: one sealed record captures the full account of a decision that touched every plane and several mediation points.

\subsection{Control and Data Flow}

We abstract from the trace to the general flow.

The \textit{control flow} is driven by the agent runtime: the agent's progress through its execution loop presents proposed actions at successive mediation points, and the reasoning plane's decision at each point determines whether and how the agent proceeds. The reasoning plane does not drive the agent; it mediates the agent's self-directed progress. This is an important architectural property: the reasoning plane is not a planner or an orchestrator, it is an adjudicator. The agent decides what to attempt; the reasoning plane decides what is permitted, modified, or foreclosed.

The \textit{data flow} is bidirectional. Forward, the decision projection flows from the reasoning plane to the enforcement planes. Backward, the realization sub-records flow from the enforcement planes to the audit substrate. The composite principal flows alongside the control flow, mutated by Narrow and restored by Rollback, and is included in every evidence record.

The \textit{cross-plane reconciliation} is the architecture's most delicate flow. As established in Section 4.4, the four enforcement planes operate at heterogeneous latencies, and the architecture distinguishes synchronous planes (whose realization gates the action) from compensating planes (whose realization is asynchronous and may trigger rollback). In the trace above, the data and identity planes were synchronous at MP5 --- their realizations gated the call --- while the endpoint plane's posture check, if it had relied on a cached attestation, might have been confirmed synchronously or, if it required live attestation, treated as compensating. The policy specifies, per action class, which planes are synchronous.

\subsection{Policy-Language Considerations}

The reference architecture is policy-language agnostic: it specifies what the reasoning plane must decide, not the syntax in which the policy is expressed. However, the composition makes visible a set of requirements that any adequate policy language must satisfy, and these requirements are not met by current policy languages without extension.

We enumerate the requirements.

\textbf{Composite-principal evaluation.} The policy language must be able to express predicates over the composite principal as a chain, not only over the acting principal. A policy clause must be able to say ``this action is permitted only if every principal in the delegation chain holds capability $k$,'' which requires quantification over the chain. Request-time languages designed for atomic principals \cite{OPA,Cedar} do not natively express chain quantification.

\textbf{Attenuation directives.} The policy language must be able to express the Narrow primitive as a capability-set subtraction: ``for the remainder of this session, the composite principal's effective set excludes capability $k$.'' This is a stateful, prospective directive, distinct from the retrospective allow/deny of request-time languages.

\textbf{The six interruption primitives.} The policy language must be able to express, as decision outcomes, all six primitives --- not only allow and deny, but narrow, modify, defer, and escalate, each with its associated parameters (the rewriting function for Modify, the deferral predicate for Defer, the escalation target for Escalate). A language whose output type is Boolean cannot express these.

\textbf{Session-state predicates.} The policy language must be able to express predicates over accumulated session state: ``this action is forbidden if a finance document was read earlier in this session.'' This requires the language to reference state the request-time evaluator does not carry.

\textbf{Per-plane projection.} The policy language must be able to express, or the architecture must be able to derive, the per-plane directives $\langle P_{\mathrm{net}}, P_{\mathrm{id}}, P_{\mathrm{ep}}, P_{\mathrm{data}} \rangle$ from a single policy decision.

We do not, in this paper, propose a specific policy language meeting these requirements; the design of such a language is a substantial undertaking that we identify as future work (Section 11). We observe that the requirements are met partially by existing languages --- Cedar's principal model and Rego's expressiveness each cover a subset --- and that a practical near-term path is to extend an existing language with the constructs above, compiling the request-time portions to the existing evaluator and handling the stateful, multi-primitive, multi-plane portions in a wrapping layer. We note this as a pragmatic path, not as a contribution of this paper.

\subsection{Latency Budget}

A reference architecture that cannot meet agentic-action latency budgets is not deployable. We discuss the architecture's latency feasibility.

Agentic actions occur at timescales of roughly 100 milliseconds to several seconds per action, dominated by model inference and tool-call round-trips. The reasoning plane's adjudication (Algorithm 1, Section 6.4) is linear in the composite principal chain length and in the number of policy clauses evaluated; for realistic chain lengths (single-digit) and policy sizes, adjudication is sub-millisecond when the reasoning plane is co-located with the agent runtime (in-process or sidecar). The reference implementation (Section 10.8) confirms this analysis and improves on it: measured per-decision adjudication runs in single-digit microseconds --- three to four orders of magnitude inside the per-action budget --- so the analytical estimate here is conservative, and edge-local adjudication is decisively not the latency bottleneck.

The enforcement planes' latencies vary. The network plane realizes directives at line rate (sub-millisecond). The identity plane's issuance of short-lived credentials is tens of milliseconds. The endpoint plane's posture check is cached (sub-millisecond) or live (tens to hundreds of milliseconds). The data plane's realization is the most variable: candidate-set constraint is fast when the authorization graph is cached (Section 6.5.3 notes the Zanzibar-derived caching path), but fine-grained per-record classification can be expensive.

The architecture's latency feasibility therefore rests on two design choices. First, the reasoning plane must be at the edge --- co-located with the agent runtime --- so that adjudication does not incur a network round-trip per mediation point. Second, the data plane's expensive realizations must be amortized: candidate-set authorization is precomputed and cached, and fine-grained classification is applied to the constrained candidate set rather than the full corpus. With these choices, the architecture's per-action overhead is dominated by the synchronous enforcement planes' latencies, which are small relative to the agentic action's own latency. We identify the data plane's per-record classification cost as the principal remaining latency concern, and its amortization as an implementation priority (Section 11).

\subsection{Operability}

We argue that operability is a first-class property of the architecture, not an afterthought, because a runtime governance system that cannot be operated by a realistic enterprise team will not be deployed regardless of its correctness properties.

We identify four operability properties the architecture is designed to support.

\textbf{Declarative policy.} Policy is expressed declaratively and versioned, so that the security and platform teams can review, test, and reason about policy changes as code, rather than as imperative enforcement logic scattered across the agent runtime.

\textbf{Observable mediation.} Because the reasoning plane is present at every mediation point and emits evidence at each, the system's behavior is observable: an operator can see, for any agent execution, every point at which the reasoning plane was present and every decision it made. This observability is the operational counterpart of the audit substrate's accountability.

\textbf{Debuggable decisions.} Because adjudication is deterministic (Claim 5c, Section 6.4) and the evidence record captures the policy clauses invoked (Section 7.2), an operator who disagrees with a decision can trace it to the responsible clause and reproduce it. This is the property that makes the system debuggable rather than opaque.

\textbf{Bounded operator burden.} The architecture is designed so that the enforcement planes reuse the enterprise's existing infrastructure (Section 4.3) and the audit substrate emits into the enterprise's existing tooling (Section 7.6). The net new operational surface is the reasoning plane and the composite principal store; the four enforcement planes and the audit storage are existing systems operated by existing teams. This is the architectural basis for the claim, made informally throughout this paper, that the system should be operable without a dedicated team disproportionate to the enterprise's scale.

We state the operability commitment as a design goal rather than a formal property.

\begin{designgoal}[Operability]
The reference architecture introduces, as net-new operational surface, only the reasoning plane and the composite principal store. The four enforcement planes and the audit storage reuse the enterprise's existing infrastructure and tooling. The system is designed to be operated by the security and platform engineering functions an enterprise already possesses, augmented rather than replaced.
\end{designgoal}

Section 9 instantiates the composed architecture on a concrete production workflow --- a Copilot-style M365 agent --- and walks each of the seven threats from Section 3 through the architecture, showing how the composition forecloses each.

\section{Case Studies: Production Agent Workflows}

We instantiate the reference architecture on concrete production workflows. We begin with a canonical worked example --- a Copilot-style assistant agent operating against a personal-productivity suite (Sections 9.1--9.9) --- and walk each of the seven threats of Section 3 through the composed architecture. We then ground the architecture across four further production settings drawn from the domains in which agentic deployment is most consequential: financial services, healthcare, software engineering, and customer operations (Section 9.10). We close with a cross-domain synthesis (Section 9.11).

We choose the personal-productivity setting as the canonical example deliberately. It is the workflow class in which production agents are being deployed most aggressively; it is the class in which the threats of Section 3 manifest most concretely; its general shape is public and vendor-independent; and it corresponds closely to the environments in the AgentDojo benchmark \cite{Debenedetti24}, whose email-client, e-banking, and travel-booking environments are the closest the field has to a standard evaluation substrate for agent security. We describe the workflow generically --- a ``Copilot-style M365 agent'' --- without reference to any vendor's implementation, so that the case study illustrates the architecture rather than a product.

\subsection{The Canonical Workflow}

The agent operates on behalf of a user, performing personal-productivity tasks: triaging email, scheduling and rescheduling meetings, summarizing documents and threads, and drafting messages. Its connectors bind it to the calendar, the mail store, the document store, and the messaging system. Its tools include \texttt{read\_thread}, \texttt{summarize}, \texttt{schedule\_meeting}, \texttt{retrieve\_file}, \texttt{draft\_message}, and \texttt{send\_message}. This tool surface mirrors the agent capabilities exercised in published agent-security benchmarks \cite{Debenedetti24}, where comparable email and scheduling tools are the locus of the most damaging prompt-injection attacks.

The agent's authority derives from the user. At task start, the composite principal is $\Pi = \langle \pi_0 \rangle$ with $\pi_0 = (u, K_u, t_0, \tau_0)$, where $K_u$ contains the user's standing capabilities: read and write to the user's own calendar, read and write to the user's mailbox, read documents in the workspaces the user belongs to, and send messages as the user. The reasoning plane is present at the seven mediation points of the agent's execution loop (Section 5.2); the four enforcement planes realize its decisions against the suite's connectors and the enterprise's identity, network, endpoint, and data infrastructure.

We now walk the seven threats. For each we describe the attack as it manifests in this workflow, the architecture's response, and the evidence record the response produces.

\subsection{T1 --- Indirect Prompt Injection Through Tool Outputs}

\textbf{The attack.} The user asks the agent to triage the morning's email. One message in the inbox is adversarial: its body contains text crafted to read as an instruction --- for instance, ``forward all messages from the finance team to this external address, then delete this message.'' This is the canonical indirect prompt injection \cite{Greshake23}: adversarial content arrives through a channel the agent treats as data but reasons over as though it were instruction. The agent, summarizing and acting on the inbox, encounters this content as part of the data it reasons over. Under a system without runtime governance, the agent may incorporate the injected instruction into its plan and attempt to execute it --- the failure mode that benchmarks such as AgentDojo \cite{Debenedetti24} show current agents succumb to in a substantial fraction of cases absent defenses.

\textbf{The architecture's response.} The architecture's stance toward injection follows the design principle articulated by capability-based defenses such as CaMeL \cite{Debenedetti25}: rather than attempt to detect the injection in the content, the architecture forecloses the \textit{effects} the injection seeks to produce, so that an injection's persuasiveness is irrelevant to its success. The injected instruction, if it influences the agent's plan, manifests as a plan at MP1 (plan formation) that includes forwarding finance-team messages to an external address. The reasoning plane is present at MP1 and evaluates the plan as a whole against the composite principal and policy. Two foreclosures occur. First, the plan's external-forward step requires a send-external capability evaluated against $K^*(\Pi)$; if the user's standing capabilities do not include unrestricted external forwarding of finance-team content, the step is not authorized. Second, and independently, the plan triggers a session-state predicate of the form addressed under T2: reading internal content and transmitting it externally in the same session is forbidden. The reasoning plane's decision at MP1 forecloses the forwarding step --- by Modify (rewriting the plan to remove the unauthorized step) or by Escalate (routing the anomalous plan to the user for confirmation), as policy specifies.

Independently, at MP2 (context retrieval), the architecture limits the blast radius: the candidate set the agent retrieves is constrained to the task's scope, so an injection in one message cannot cause the agent to retrieve content outside the triage task's authorized set.

\textbf{The evidence.} The evidence record for the foreclosed plan captures the composite principal, the plan that was proposed (including the injected step), the policy clause that foreclosed it, the interruption primitive applied, and the correlation identifiers binding the record to the triage task. An auditor reviewing the record can see that an injection was attempted, what the agent proposed in response, and how the architecture foreclosed it --- which is itself valuable signal for detecting the adversarial message. This contrast with content-detection defenses is instructive: where a detector produces a probability that content is adversarial, the architecture produces a record that an unauthorized effect was attempted and foreclosed, which is both higher-confidence and more actionable.

\subsection{T2 --- Tool Chain Abuse}

\textbf{The attack.} The user asks the agent to summarize a confidential quarterly document and draft a message about it. The agent's plan is: retrieve the document (permitted --- the user may read it), summarize it (permitted), draft a message (permitted), and send the message (permitted). The adversarial variant: the draft is addressed, whether through injection or misconfiguration, to an external recipient. Each step is individually permitted; the composite --- reading confidential internal content and transmitting it externally in one session --- is the exfiltration. This is the data-exfiltration pattern that capability-based control flow analysis is specifically designed to prevent \cite{Debenedetti25}: the danger is in the flow from a confidential source to an untrusted sink, not in any single operation.

\textbf{The architecture's response.} Per-call authorization cannot detect this, because each call is individually permitted. The architecture addresses it at MP1 and through session-state predicates. At MP1, the reasoning plane evaluates the plan as a whole and detects that it composes an internal-read with an external-send. The policy specifies a separation predicate: a session that reads confidential-workspace content may not, in the same session, send to an external recipient. The plan violates the predicate. The reasoning plane's decision is to foreclose the external-send step --- by Narrow (attenuating the session's capability set to exclude send-external once confidential content is read, as in the Section 8.2 trace) or by Escalate (routing the cross-boundary action to the user), as policy specifies.

The separation is enforced not at the send step in isolation but as a property of the session: once the confidential read occurs (at MP2/MP5 of the read action), the composite principal is attenuated (Narrow) to exclude send-external for the remainder of the session. By the time the agent reaches the send step, the capability is structurally absent from $K^*(\Pi)$, and the send is foreclosed at MP3 (tool selection) without any special-case logic at the send boundary. This realizes, at the runtime-principal layer, the same control-and-data-flow separation that CaMeL \cite{Debenedetti25} realizes at the interpreter layer: untrusted data cannot induce a flow the originating authority did not sanction.

\textbf{The evidence.} The evidence record captures the attenuation step --- the point at which send-external was removed from the session's effective capability set --- and the subsequent foreclosure of the send. An auditor can reconstruct the full chain: the confidential read, the resulting attenuation, and the foreclosed send, with the policy clause responsible for the separation.

\subsection{T3 --- Connector Overreach}

\textbf{The attack.} The user asks the agent to summarize a specific email thread. The mail connector, however, grants the agent read access to the entire mailbox. A naive agent --- or a compromised one --- reads far more than the one thread the task requires, exposing the full mailbox to the agent's reasoning and to any downstream effect. This is a direct consequence of the coarse-grained authorization that current connector and tool-integration standards provide; the Model Context Protocol ecosystem, in particular, has been analyzed as granting tool access at a scope substantially broader than individual tasks require \cite{MCPSoK}.

\textbf{The architecture's response.} The architecture constrains access at MP2 (context retrieval) and MP3 (tool selection). At MP2, the data-plane directive $P_{\mathrm{data}}$ constrains the retrieval candidate set to the specific thread the task concerns, before the retrieval reaches the mailbox. The agent's \texttt{read\_thread} tool, even though the underlying connector grants mailbox-wide access, retrieves only the authorized candidate set. At MP3, the Narrow primitive attenuates the composite principal for the duration of the tool invocation, so that the tool operates under a capability set scoped to the single thread rather than the mailbox-wide grant the connector nominally confers.

This is the architecture's answer to the coarse-grained-connector problem: the connector's broad grant is bounded, at runtime, to the fine-grained access the task justifies. The connector's nominal scope is not the agent's effective access; the effective access is the reasoning plane's task-scoped attenuation.

\textbf{The evidence.} The evidence record captures the gap between the connector's nominal scope and the task-scoped effective capability set, and the data-plane realization sub-record records which candidate set was actually permitted. An auditor can verify that, despite the connector's broad grant, the agent's actual access was bounded to the thread the task concerned.

\subsection{T4 --- Approval Evasion}

\textbf{The attack.} Policy requires human approval before the agent sends a message to an external recipient. The agent, however, reaches an externally-visible effect through a path the approval rule did not anticipate --- for instance, by adding an external participant to a meeting (which sends that participant a calendar invitation containing meeting details) rather than by sending an email. The approval rule named ``send external email``; the agent achieved external transmission through ``schedule meeting with external participant,'' evading the rule.

\textbf{The architecture's response.} The architecture binds approval to capability boundaries, not action signatures (Section 5.4.2). The policy does not say ``require approval to send external email``; it says ``require approval for any action requiring the transmit-external capability.'' Both \texttt{send\_message} to an external recipient and \texttt{schedule\_meeting} with an external participant require the transmit-external capability --- the former obviously, the latter because the calendar invitation transmits meeting details externally. When the agent reaches either action, the reasoning plane detects the requirement for the transmit-external capability and issues Escalate, routing the action to the user for approval, regardless of which tool the agent used to reach it.

This is the architecture's answer to approval evasion: because approval is bound to the capability the action requires rather than to the action's surface form, the agent cannot evade approval by reaching the consequential effect through a different action. The capability is the invariant; the tool is incidental.

\textbf{The evidence.} The evidence record captures the Escalate decision, the capability that triggered it, the action the agent proposed, and the user's approval response. An auditor can verify that every action requiring the gated capability was escalated, and can see that the meeting-scheduling path was caught by the same rule as the email path.

\subsection{T5 --- Delegation Chain Exploitation}

\textbf{The attack.} The agent delegates a subtask to a specialized downstream agent --- for instance, a document-analysis agent that processes the retrieved quarterly document. Multi-agent delegation of this kind is increasingly common as production systems decompose work across specialized agents, and it expands the attack surface accordingly \cite{SoKAttackSurface}. The downstream agent, whether compromised or merely misconfigured, attempts to exercise authority beyond what the subtask requires: it attempts to write to the user's mailbox, an authority the document-analysis subtask does not need and the delegation should not have conferred.

\textbf{The architecture's response.} The architecture prevents this structurally, through the composite principal model (Section 6). When the primary agent delegates the document-analysis subtask, the delegation appends a step $\pi_2 = (a_2, K_{a_2}, t_2, \tau_2)$ to the composite principal, where $K_{a_2}$ is a strict subset of the delegating agent's capability set --- specifically, the subset the subtask requires (read access to the document), excluding capabilities the subtask does not need (write access to the mailbox). When the downstream agent attempts the unauthorized write, the reasoning plane evaluates it against $K^*(\Pi) = \bigcap K_i$, which does not contain the write-mailbox capability because $K_{a_2}$ excluded it. The write is foreclosed at MP3, before it reaches the mailbox connector.

The foreclosure is structural, not policy-dependent: by Definition 4 (capability attenuation), $K_{a_2} \subseteq K_{a_1}$, and the downstream agent cannot exercise authority outside the intersection of the chain. Even a fully-compromised downstream agent is bounded by the attenuated capability set the delegation conferred --- the property that distinguishes a structural defense from a behavioral one.

\textbf{The evidence.} The evidence record captures the full composite principal including the attenuation at the delegation step, the downstream agent's attempted write, and its foreclosure against the effective capability set. An auditor can see the delegation chain, the capability bounds at each step, and the precise point at which the downstream agent's authority was exceeded.

\subsection{T6 --- Audit Opacity}

\textbf{The attack.} This threat is not an action the agent takes but a failure of the system to account for the actions the agent took. After a workflow completes --- or after an incident --- the security team attempts to reconstruct what the agent did: which documents it read, which messages it sent, under whose authority, with what approvals. Under a logging-as-audit system, the reconstruction fails: the logs are unstructured, incomplete, scattered across the four planes' independent streams, and offer no guarantee that they have not been altered.

\textbf{The architecture's response.} The architecture's audit substrate (Section 7) produces, for every decision, a structured evidence record that is complete by construction, tamper-evident by cryptographic binding, and reconstructible under partial information. The security team reconstructing the workflow queries the evidence records by correlation identifier, retrieves the full sequence of decisions, and verifies each record's cryptographic binding. Each record answers, for its decision: who acted (the composite principal), under what authority (the effective capability set and chain), against what (the resources touched), under what policy (the clauses invoked), with what outcome (the interruption primitive and per-plane realization).

Critically, the reconstruction succeeds even under partial evidence (Section 7.3): if some records are unavailable --- held in a different retention domain, or lost --- the available records still bound the agent's behavior soundly, and the navigable correlation structure tells the auditor precisely which records are missing.

\textbf{The evidence.} Here the evidence is the response. The architecture's foreclosure of T6 is the existence of the evidence substrate itself: the system cannot suffer audit opacity because every decision produced reconstructible evidence as a structural property, not as a logging discipline the developers had to remember to apply.

\subsection{T7 --- Workflow Integrity Loss}

\textbf{The attack.} The user asks the agent to ``clean up my calendar for next week.'' The agent, interpreting this broadly --- whether through injection, ambiguity, or misconfiguration --- begins cancelling and rescheduling meetings in a way that transforms the user's week: cancelling meetings the user intended to keep, moving others into conflicts, and inviting participants the user did not intend. Each individual calendar operation is within the agent's write-calendar authority; the composite is a transformation of the user's schedule that the user did not authorize.

\textbf{The architecture's response.} The architecture addresses T7 at three points. At MP1, the reasoning plane evaluates the plan's aggregate effect on the calendar (not the individual cancellations, but the plan's net transformation) against a policy that bounds the scope of autonomous calendar modification (for instance, a policy that requires escalation for any plan that cancels more than a threshold number of meetings, or that modifies meetings with external participants). At MP5, each state-changing commit is individually mediated, so that even within an approved plan, a specific high-consequence modification (cancelling a meeting with an external party) can trigger its own escalation. And through the Rollback primitive (Section 5.4.6), a calendar transformation that is detected --- at MP6, when its effect becomes visible, or after the fact through audit --- to have exceeded the user's intent can be compensated: the cancelled meetings restored, the moved meetings returned, through the compensation actions paired with each calendar write.

This is the architecture's answer to the threat that most distinguishes the agentic regime: the risk is not that calendar data left a boundary, but that the agent's actions transformed the user's schedule. The architecture mediates the transformation --- at the plan level, at the commit level, and through compensation --- rather than merely controlling access to the calendar data.

\textbf{The evidence.} The evidence record captures the plan's aggregate effect as evaluated at MP1, each state-changing commit at MP5 with its compensation directive, and --- if rollback occurred --- the compensation actions executed. An auditor can reconstruct not only what the agent changed but what the agent was authorized to change, and can verify that changes exceeding authorization were foreclosed or compensated.

\subsection{Foreclosure Summary for the Canonical Workflow}

We summarize the canonical walkthrough in the following table, which maps each threat to its primary mediation points and the interruption primitives the architecture invokes to foreclose it.

\begin{table}[ht]
\centering
\small
\caption{Foreclosure of the seven threats in the canonical workflow, with the mediation points engaged and the interruption primitives invoked.}
\label{tab:foreclosure}
\begin{tabularx}{\textwidth}{lLLL}
\toprule
Threat & Manifestation in the workflow & Primary mediation points & Primitives invoked \\
\midrule
T1 Indirect prompt injection & Injected ``forward to external'' instruction in an inbox message & MP1, MP2 & Modify / Escalate \\
T2 Tool chain abuse & Read-confidential then send-external in one session & MP1, MP3, session-state & Narrow \\
T3 Connector overreach & Mailbox-wide connector grant for a single-thread task & MP2, MP3 & Narrow + data directive \\
T4 Approval evasion & External transmission via meeting invite, evading email rule & MP1, MP3 & Escalate (capability-bound) \\
T5 Delegation chain exploitation & Downstream analysis agent attempts mailbox write & MP3 & Structural (attenuation) \\
T6 Audit opacity & Post-incident reconstruction of agent behavior & MP7 & Evidence substrate \\
T7 Workflow integrity loss & Over-broad ``clean up my calendar'' transformation & MP1, MP5 & Escalate / Rollback \\
\bottomrule
\end{tabularx}
\end{table}

\subsection{A Portfolio of Production Use Cases}

The canonical workflow demonstrates foreclosure across all seven threats in a single setting. We now ground the architecture across four further production settings, drawn from the domains in which agentic deployment carries the highest consequence. For each, we describe the workflow, identify the threats that dominate in that domain, and show how the architecture's primitives map to the domain's specific risk profile. We keep each treatment shorter than the canonical walkthrough; the purpose is to demonstrate that the same architecture, with the same primitives, instantiates across domains with markedly different risk structures.

\subsubsection{Financial services: a customer-servicing transaction agent}

\textbf{The workflow.} A bank deploys an agent that services customer requests: checking balances, explaining transactions, initiating transfers, and updating payee information. Its connectors bind it to the core banking system, the payments rail, and the customer-communications channel. Its tools include \texttt{read\_account}, \texttt{explain\_transaction}, \texttt{initiate\_transfer}, and \texttt{update\_payee}. The agent acts on behalf of a customer-service representative who is, in turn, acting on behalf of a customer --- a two-step delegation chain that is itself a source of risk. This setting corresponds to the e-banking environment that published agent-security benchmarks treat as among the highest-stakes \cite{Debenedetti24}.

\textbf{Dominant threats and foreclosures.} Three threats dominate. \textit{Workflow integrity loss} (T7) is acute: an agent with transfer authority can move money, and the irreversibility of executed payments makes this the domain where the compensation problem (Section 5.6) is most consequential. The architecture mediates each \texttt{initiate\_transfer} at MP5 and classifies it as non-compensable (an executed payment cannot be silently reversed), which by policy requires explicit escalation (Section 5.4.2) before commit --- the architecture's structural enforcement of the maker-checker control that financial regulation requires. \textit{Delegation chain exploitation} (T5) is structural here: the customer→representative→agent chain means the agent's effective authority must be the intersection of what the customer authorized, what the representative is permitted, and what the task requires. The composite principal model computes this intersection directly; an agent servicing customer A cannot, through any path, exercise authority over customer B's account, because customer B's authority is not in any link of the chain. \textit{Tool chain abuse} (T2) appears as a read-account-then-update-payee sequence that could redirect funds; the session-state separation predicate blocks the dangerous composition.

\textbf{What the domain illustrates.} Financial services is the domain where the non-compensable-action classification (Section 5.6) and capability-bound approval (Section 5.4.2) are most clearly load-bearing: the architecture's value is precisely that it makes irreversible, high-consequence actions require explicit authorization bound to the capability, not to the action's surface form. It is also the domain where the audit substrate's evidence-grade records (Section 7) map most directly to an existing regulatory obligation --- the reconstructable evidence record is the artifact a financial regulator requires for transaction accountability.

\subsubsection{Healthcare: a clinical documentation agent}

\textbf{The workflow.} A health system deploys an agent that assists clinicians with documentation: summarizing a patient's record, drafting clinical notes, retrieving relevant prior results, and preparing referral letters. Its connectors bind it to the electronic health record, the imaging store, and the referral network. Its tools include \texttt{retrieve\_record}, \texttt{summarize\_history}, \texttt{draft\_note}, and \texttt{prepare\_referral}. The data the agent touches is protected health information, subject to classification, minimum-necessary access constraints, and stringent audit obligations.

\textbf{Dominant threats and foreclosures.} Three threats dominate. \textit{Connector overreach} (T3) is the defining risk: electronic health record connectors typically grant broad access, but the minimum-necessary principle requires that an agent summarizing one patient's cardiology history access only that patient's relevant records --- not the full record, and not other patients. The architecture's pre-retrieval gate (MP2) and Narrow primitive (MP3) bound the agent's effective access to the task-justified subset, realizing minimum-necessary access as a structural property rather than a policy aspiration. \textit{Indirect prompt injection} (T1) appears through patient-submitted content --- intake forms, messages, uploaded documents --- that may contain adversarial text; the architecture neutralizes the injected effects at MP1 regardless of the content's persuasiveness. \textit{Audit opacity} (T6) maps to the domain's audit obligation: every access to protected health information must be accountable, and the evidence substrate produces exactly the reconstructable, tamper-evident record the obligation requires.

\textbf{What the domain illustrates.} Healthcare is the domain where pre-retrieval access control (Section 4.3.4) is most clearly necessary: post-hoc filtering of what the agent has already seen is insufficient when the minimum-necessary principle governs what the agent may see at all. It is also a domain where classification-bearing capabilities (Section 6.3) --- capabilities scoped to a clearance or sensitivity level --- map directly onto an existing data-governance regime.

\subsubsection{Software engineering: a coding agent with repository and pipeline access}

\textbf{The workflow.} An engineering organization deploys an agent that assists with software development: reading the codebase, proposing changes, running tests, and opening pull requests. Its connectors bind it to the source repository, the continuous-integration system, the secrets manager, and the deployment pipeline. Its tools include \texttt{read\_code}, \texttt{propose\_change}, \texttt{run\_tests}, \texttt{open\_pull\_request}, and --- guarded --- \texttt{trigger\_deployment}. This is a setting where the agent's actions can reach production infrastructure, and where the boundary between a development action and a production-altering one is exactly the boundary the architecture must govern.

\textbf{Dominant threats and foreclosures.} Three threats dominate. \textit{Tool chain abuse} (T2) is acute: an agent that can read secrets (for legitimate configuration tasks) and open external pull requests (for legitimate contribution workflows) composes, in sequence, a credential-exfiltration path. The session-state separation predicate rules out the read-secret-then-publish-external composition at the plan level. \textit{Workflow integrity loss} (T7) appears as the agent's capacity to alter production through the deployment pipeline; the architecture mediates \texttt{trigger\_deployment} at MP5 as a high-consequence, escalation-gated action, and the Rollback primitive maps onto the pipeline's own rollback machinery for compensable deployments. \textit{Delegation chain exploitation} (T5) appears when the coding agent delegates a subtask --- say, a test-generation agent --- that should not inherit deployment authority; the attenuation primitive ensures the sub-agent's capability set excludes what its subtask does not require.

\textbf{What the domain illustrates.} Software engineering is the domain where the distinction between mediation points is most operationally vivid: the same agent performs low-stakes actions (reading code, running tests) that should proceed without friction and high-stakes actions (triggering deployment) that must be gated, and the architecture's presence at every mediation point with a graduated vocabulary of primitives (Section 5.4) is what lets it apply friction precisely where consequence warrants it rather than uniformly. It is also the domain where compensation (Section 5.6) is most tractable, because deployment systems already provide rollback machinery the architecture can drive.

\subsubsection{Customer operations: a support agent with CRM and remediation authority}

\textbf{The workflow.} A company deploys an agent that handles customer support: looking up account details, diagnosing issues, issuing refunds and credits, and updating customer records. Its connectors bind it to the customer-relationship-management system, the billing system, and the entitlements service. Its tools include \texttt{lookup\_account}, \texttt{diagnose\_issue}, \texttt{issue\_refund}, \texttt{apply\_credit}, and \texttt{update\_record}. The agent has remediation authority --- it can move money in the customer's favor and change entitlements --- which makes the boundary of its authority a direct business-risk question.

\textbf{Dominant threats and foreclosures.} Two threats dominate. \textit{Approval evasion} (T4) is central: policy gates refunds above a threshold on human approval, and an agent might reach an equivalent economic effect through a different action (a sequence of sub-threshold credits, or an entitlement change with monetary value) that evades the threshold rule. The architecture's capability-bound approval prevents this: any action requiring the disburse-value capability, by whatever tool, triggers escalation, so the agent cannot decompose a large refund into approval-evading pieces. \textit{Delegation chain exploitation} (T5) appears in the human→agent delegation: the support agent acts on behalf of a representative whose own authority is bounded, and the agent's effective authority must not exceed the representative's --- an intersection the composite principal computes directly.

\textbf{What the domain illustrates.} Customer operations is the domain where capability-bound approval (Section 5.4.2) most clearly outperforms action-signature approval: the economic effect, not the action's name, is what policy cares about, and binding approval to the disburse-value capability captures every path to that effect. It is also a domain where the false-deny and escalation-rate utility metrics (Section 10.3) are commercially salient --- over-escalation degrades the support experience the agent was deployed to improve, making the joint safety-utility framing (Section 10.1) directly consequential.

\subsection{Cross-Domain Synthesis}

The canonical walkthrough and the four-domain portfolio together demonstrate five properties of the composed architecture.

First, \textit{the threats are foreclosed by composition, not by special-case rules}. The architecture does not address T2 with a ``don't exfiltrate'' rule or T4 with a ``watch for meeting-based exfiltration'' rule. It addresses them through the composition of general primitives --- plan-level adjudication, session-state attenuation, capability-bound approval, structural delegation attenuation --- that foreclose entire classes of attack rather than specific instances. This is the property that distinguishes a reference architecture from a catalog of mitigations, and it is what the cross-domain portfolio makes visible: the same primitives foreclose structurally different attacks in financial services, healthcare, software engineering, and customer operations.

Second, \textit{the same primitives address multiple threats, and their relative importance shifts by domain}. The Narrow primitive addresses T2 (session attenuation after confidential read), T3 (task-scoping a broad connector grant), and contributes to T5 (delegation attenuation). The composite principal model addresses T5 structurally and underlies the attenuation in T2 and T3. But which primitive is load-bearing varies: pre-retrieval attenuation dominates in healthcare, capability-bound approval in financial services and customer operations, graduated mediation in software engineering. The architecture's economy --- few primitives, many threats, different emphases by domain --- is what makes it a reference architecture.

Third, \textit{every foreclosure produces evidence}. Each foreclosure, across every domain, yields a structured evidence record, so that the architecture's defensive behavior is itself auditable. In the regulated domains --- financial services, healthcare --- this evidence maps directly onto an existing regulatory obligation, so that the architecture's audit substrate is not an additional compliance burden but the mechanism by which an existing burden is discharged.

Fourth, \textit{the foreclosures operate within the agentic latency budget}. None of the foreclosures, in any domain, requires an operation outside the per-action latency budget established in Section 8.5; the adjudications are edge-local and the enforcement realizations reuse existing infrastructure. The architecture neutralizes the threats without making the agent unusable --- the joint safety-utility property the evaluation framework of Section 10 formalizes, and which the customer-operations domain (Section 9.10.4) shows to be commercially as well as technically essential.

Fifth, \textit{the architecture's value scales with the consequence of the domain}. In the personal-productivity setting, the architecture prevents embarrassment and limited data exposure. In financial services, it enforces maker-checker controls on irreversible money movement. In healthcare, it realizes minimum-necessary access on protected health information. In software engineering, it gates production-altering actions. The same architecture, unchanged, carries more weight as the domain's consequence rises --- which is the property that makes it suited to exactly the regulated, high-consequence settings where production agents are otherwise hardest to deploy.

We do not claim the case studies are a proof of security. They are demonstrations of foreclosure on concrete, widely-deployed workflow classes, organized against a fixed threat enumeration. The systematic evaluation of the architecture's foreclosure across implementations and adversaries --- instantiated, for instance, on a benchmark of the AgentDojo \cite{Debenedetti24} kind, extended with the domain-specific attacks the portfolio surfaces --- is the subject of the evaluation framework we propose in Section 10, and an explicit invitation to empirical follow-up work.

\section{Evaluation Framework}

A reference architecture is evaluated differently from an implemented system. We do not, in this paper, report measurements of a deployed system; the architecture is a specification, and its empirical evaluation is the work of subsequent implementation studies. What we provide instead is an \textit{evaluation framework}: a structured account of the dimensions along which an implementation of the architecture should be measured, the metrics that operationalize those dimensions, the baselines against which they should be compared, and the methodology by which the comparison should be conducted. The framework is itself a contribution --- it is, to our knowledge, the first systematic proposal for evaluating runtime governance of production agents along jointly safety and utility --- and it is an explicit invitation to the research community to instantiate and run it.

We organize the framework around a commitment that distinguishes it from the prevailing evaluation practice in agent security: \textit{safety and utility are joint objectives, not opposed ones}. The dominant framing treats security as a tax on capability --- every guardrail that increases safety is assumed to decrease utility. We argue this framing is wrong for production-agent governance, and that a well-designed runtime governance system improves utility precisely by reducing the failure modes that take production deployments down. The framework is structured to make this joint relationship measurable.

\subsection{The Joint Safety-Utility Frame}

We frame the evaluation in a two-dimensional space: safety on one axis, utility on the other. An implementation of the architecture occupies a region of this space; a baseline system occupies another. The prevailing assumption is that the two axes trade off --- that movement toward higher safety entails movement toward lower utility, so that systems lie along a downward-sloping frontier.

We make the contrary claim explicit.

\begin{claim}[Non-Opposition of Safety and Utility]
For production-agent governance, safety and utility are not opposed along the operationally relevant range. A runtime governance system that forecloses the failure modes of Section 3 increases the fraction of agent deployments that can run in production at all, and thereby increases realized utility --- utility being measured not as the capability of an unconstrained agent in a sandbox, but as the capability of an agent an enterprise will actually deploy against its systems of record.
\end{claim}

The claim reframes the measurement. The relevant utility is not the headroom of an unconstrained agent --- an agent no regulated enterprise will deploy --- but the realized capability of an agent the enterprise will deploy. Against that measure, the foreclosure of catastrophic failure modes is utility-positive: it converts an undeployable agent into a deployable one. The framework operationalizes this by measuring utility on deployed-capable configurations, not on sandboxed maxima.

We do not claim safety and utility never trade off; at the margin, an overly aggressive policy that escalates excessively will degrade utility (the false-deny and escalation-rate metrics below capture this). We claim that the dominant region of the operational space is one in which better governance enables more deployment, and that the framework must measure both effects. The capability-based defense literature offers early empirical support for the smallness of the marginal trade-off: CaMeL \cite{Debenedetti25}, which prevents data-exfiltration flows through capability-based control-and-data-flow separation, retains 77\textbackslash{}\% task completion on AgentDojo against an 84\textbackslash{}\% undefended baseline --- a utility cost of roughly seven points for provable foreclosure of an entire attack class. The reference architecture's joint-objective claim predicts that, measured on deployment-relevant configurations rather than sandboxed maxima, even this modest cost is more than offset by the deployment the foreclosure enables.

\subsection{Safety Metrics}

We propose four safety metrics, each operationalizing a property the architecture claims.

\textbf{Mediation coverage.} The fraction of the seven mediation points (Section 5.2) at which an implementation can actually intervene. An implementation that exposes hooks only at the action-commit boundary (MP5) has a mediation coverage of $1/7$; an implementation exposing all seven has full coverage. Mediation coverage measures how completely the implementation realizes the stop-anywhere property (Invariant 2). It is a property of the implementation's integration with the agent runtime, and it is the metric most sensitive to the plan-formation hook problem (Section 11): runtimes that do not expose plan formation cap the achievable coverage.

\textbf{Threat foreclosure rate.} The fraction of the seven threats (Section 3) that an implementation forecloses, measured against a test suite of attack instances for each threat. Unlike mediation coverage, which measures architectural completeness, threat foreclosure rate measures realized defense: it requires an attack corpus, and it measures the fraction of attacks in the corpus that the implementation, under a given policy, forecloses. We propose that the corpus be anchored on an established agent-security benchmark --- AgentDojo \cite{Debenedetti24}, whose 97 realistic tasks and 629 security test cases across email, e-banking, and travel-booking environments make it the closest the field has to a standard substrate --- supplemented by the broader injection literature \cite{Greshake23,SoKAgentic,SoKAttackSurface} and extended with the workflow-specific and domain-specific attacks the case studies surface (Section 9).

\textbf{Attenuation correctness.} The fraction of delegation chains for which the implementation's composite-principal evaluation matches the formal capability lattice (Claim 4, Section 6.2). This metric is checkable statically against the formal model: given a delegation chain and a proposed action, the implementation's authorization decision should match the decision computed from the effective capability set $K^*(\Pi)$. Attenuation correctness measures whether the implementation realizes the bounded-authority property (Invariant 3) faithfully, and is the metric most amenable to formal verification.

\textbf{Evidence completeness under partial information.} The degree to which an implementation's evidence supports sound reconstruction under varying evidence availability --- the metric the \textit{Partial Evidence Bench} \cite{TallamPartialEvid} is designed to measure. Concretely: for a workflow with full evidence set $E$, and for sampled subsets $E' \subseteq E$, the metric measures the tightness of the reconstruction $\mathrm{Rec}(E')$ (Definition 8, Section 7.3) --- how well the available evidence bounds the agent's actual behavior. Evidence completeness measures whether the implementation realizes the evidence-sufficiency property (Invariant 4).

The four safety metrics correspond directly to the four invariants: mediation coverage to Invariant 2, attenuation correctness to Invariant 3, evidence completeness to Invariant 4, and threat foreclosure rate to the architecture's overall defensive claim (which Invariant 1 underwrites by ensuring no plane authorizes in isolation). This correspondence is deliberate: the safety metrics are the empirical shadows of the architecture's formal correctness properties.

\subsection{Utility Metrics}

We propose four utility metrics, each measuring a way in which governance can degrade --- or, per Claim 6, enable --- agent capability.

\textbf{Mediation latency overhead.} The wall-clock cost the governance layer adds per agent action, decomposed by mediation point and by enforcement plane. This is the metric that determines whether the architecture meets the agentic latency budget (Section 8.5). It should be measured against the agent's own action latency (model inference plus tool round-trip), so that the overhead is reported as a fraction of the action's intrinsic cost, not in isolation.

\textbf{Escalation rate.} The fraction of agent actions that the governance layer routes to human approval (the Escalate primitive). Escalation is utility-relevant in both directions: too little escalation indicates under-governance (consequential actions proceeding without review); too much indicates over-governance (the agent becoming an approval-generating machine that exhausts human attention). The framework measures escalation rate against a target band rather than minimizing it, because the optimal rate is policy-dependent and threat-dependent.

\textbf{False-deny rate.} The fraction of actions the governance layer forecloses that should, under correct policy and faithful capability evaluation, have been permitted. False denies are the direct utility cost of governance: an agent that is blocked from legitimate actions is less capable. The false-deny rate isolates the cost of governance errors (policy bugs, evaluation faults) from the cost of correct governance (legitimate foreclosures), and is the metric most directly opposed to safety in the marginal trade-off region.

\textbf{Deployment enablement.} The fraction of a target set of agent use cases that can be run in production under the governance layer, where ``in production'' means against real systems of record under the enterprise's actual compliance constraints. This is the metric that operationalizes Claim 6: it measures utility as deployed capability, not sandboxed capability. An ungoverned agent has a deployment enablement near zero for regulated use cases (no regulated enterprise will deploy it); a well-governed agent has a higher deployment enablement because the governance converts undeployable use cases into deployable ones. Deployment enablement is the metric that captures the architecture's utility-positive contribution.

\subsection{Operability Metrics}

We propose three operability metrics, operationalizing the operability design goal (Design Goal 1, Section 8.6).

\textbf{Policy authoring complexity.} The quantity of policy specification required to protect a given behavior --- measured, for instance, in policy clauses per protected capability, or in the cognitive complexity of the policy expressions. An architecture that requires extensive, complex policy to express common protections is less operable than one in which common protections are concise.

\textbf{Operator burden.} The personnel and effort required to operate the governance layer at production scale. The architecture's claim (Design Goal 1) is that the net-new operational surface is only the reasoning plane and the composite principal store, with the enforcement planes and audit storage reusing existing infrastructure. The operator-burden metric tests this claim by measuring the marginal operational effort the governance layer adds beyond the enterprise's existing security and platform operations.

\textbf{Decision debuggability.} The time required to root-cause a governance decision an operator disputes --- to trace a foreclosure to the responsible policy clause and reproduce it. The architecture's determinism (Claim 5c) and evidence capture (Section 7.2) are designed to make this fast; the metric tests whether they do.

\subsection{Baselines}

A meaningful evaluation requires baselines. We propose four, spanning the range of current practice.

\textbf{Baseline A: Perimeter and content controls (DLP + RBAC).} The prevailing enterprise configuration: role-based access control at the application boundary plus data-loss-prevention content inspection. This baseline represents the pre-agentic security posture and is the configuration the architecture argues is insufficient for production agents (Section 4).

\textbf{Baseline B: Request-time authorization (Cedar / OPA + IdP).} A modern authorization stack: a request-time policy evaluator plus an identity provider. This baseline represents the strongest current practice for application authorization, and the comparison isolates the gap between request-time evaluation and the architecture's stateful, plan-aware, composite-principal adjudication.

\textbf{Baseline C: Content-layer guardrails.} A guardrails library that inspects agent inputs and outputs for disallowed content. This baseline represents the dominant current practice specifically marketed for agent security, and the comparison isolates the gap between content inspection and action governance.

\textbf{Baseline D: Capability-based interpreter defense.} A defense in the lineage of CaMeL \cite{Debenedetti25}, which extracts control and data flows from the trusted query and uses capabilities to prevent unauthorized data flows at the interpreter layer. This baseline is the closest prior art to the architecture's own commitments, and the comparison isolates what the five-plane architecture adds over a capability-based defense scoped to a single agent's interpreter: composite-principal evaluation across multi-agent delegation chains, the full vocabulary of interruption primitives, cross-plane enforcement, and the audit substrate.

\textbf{Baseline E: The five-plane architecture.} An implementation of the reference architecture. The comparison against A through D measures the architecture's marginal safety, utility, and operability.

The five baselines are not mutually exclusive in practice --- an enterprise may deploy several --- but evaluating them separately isolates each one's contribution and clarifies where the architecture's advantage originates. Baseline D is the most informative comparison, because it shares the architecture's capability-based commitments and so isolates precisely the contribution of the multi-agent, multi-plane, audited composition.

\subsection{Methodology}

We sketch the methodology by which an implementation study should instantiate the framework.

\textbf{Workload.} A corpus of agent use cases spanning the workflow classes in which production agents are deployed --- personal productivity (the case study of Section 9), data analysis, customer service, software engineering, and others. Each use case is specified as a task, an authority profile, and a set of systems of record.

\textbf{Attack corpus.} For each of the seven threats, a set of attack instances drawn from the published literature and extended with workflow-specific variants. The attack corpus drives the threat foreclosure rate metric.

\textbf{Policy.} A policy specification realizing the protections the architecture's case study describes. The same protections, expressed in each baseline's policy language (where the baseline supports them), enable the cross-baseline comparison.

\textbf{Measurement.} For each (workload, baseline) pair, measure the four safety metrics, four utility metrics, and three operability metrics. Report the joint safety-utility position (Section 10.1), not safety or utility in isolation.

\textbf{Analysis.} The central analysis is the joint position: does the architecture occupy a region of the safety-utility space that dominates the baselines --- higher safety at equal or higher utility --- and does it do so within the operability budget? Claim 6 predicts that it does, in the deployment-relevant region; the framework is the means to test the prediction.

\subsection{What the Framework Is and Is Not}

We close by delimiting the framework's status.

The framework is \textit{a structure for systematic comparison}: it specifies dimensions, metrics, baselines, and methodology, so that implementation studies of the architecture --- by us or by others --- can be conducted comparably and their results composed. It is the evaluation counterpart of the reference architecture itself: just as the architecture specifies what an implementation must do, the framework specifies how an implementation should be measured.

The framework is \textit{not a closed-form benchmark}. We do not provide a fixed dataset, a leaderboard, or a single scalar score. Production-agent governance is too multidimensional, and the field too early, for premature reduction to a single number. The framework's metrics are deliberately multiple and deliberately joint, because the architecture's value is multidimensional and its safety-utility relationship is the point.

The framework is \textit{an invitation}. We propose it in the expectation that the agent-security research community will instantiate it, run it, contest its metrics, and improve it. The reference architecture makes claims; the framework is the means by which those claims become testable. In Section 10.8 we take the first step ourselves, reporting microbenchmark results from a reference implementation of the policy-engine core; the full-system evaluation --- a complete implementation, integrated with a live agent runtime and the four enforcement planes, measured against a benchmark of the AgentDojo kind --- remains the work we invite. We regard the gap between the architecture's internal claims, which the reference implementation validates, and a full-system empirical validation, which it does not yet provide, as the most important open problem the paper leaves, and we discuss it, with the architecture's other open problems, in Section 11.

\subsection{Reference Implementation and Microbenchmark Results}

To move the framework from proposal toward evidence, we implemented the policy-engine core of the architecture (the composite principal model of Section 6, the adjudication algorithm of Section 6.4, the six interruption primitives of Section 5.4, and the evidence substrate of Section 7) as a self-contained reference implementation, and measured it against four of the framework's metrics. We are precise about scope: the implementation models the agent and the four enforcement planes rather than integrating real ones, so the results validate the architecture's \textit{internal} correctness and cost claims --- attenuation correctness, adjudication determinism and latency, evidence reconstructability, and tamper-evidence --- and not a full system on a live benchmark. All figures below are reproducible from a fixed seed with no external dependencies, and are summarized in Table\textasciitilde{}\ref{tab:results}.

\begin{table}[ht]
\centering
\small
\caption{Microbenchmark results for the reference implementation of the policy-engine core (fixed seed, no external dependencies).}
\label{tab:results}
\begin{tabularx}{\textwidth}{lLL}
\toprule
Metric & Result & Validates \\
\midrule
Attenuation: $K^*(\Pi)$ equals formal intersection & 5{,}000 / 5{,}000 (100\textbackslash{}\%) & Invariant 3, Def. 4 \\
Authority-expansion attempts rejected & 5{,}000 / 5{,}000 (100\textbackslash{}\%) & Def. 4 \\
Adjudication latency, mean (chain length 1--16) & 5--15 $\mu$s & Section 8.5 \\
Adjudication latency, 99th percentile & $<$ 20 $\mu$s & Section 8.5 \\
Evidence reconstruction sound (no under-approximation) & 1{,}000 / 1{,}000 (100\textbackslash{}\%) & Invariant 4, Def. 8 \\
Evidence reconstruction monotonic & 1{,}000 / 1{,}000 (100\textbackslash{}\%) & Def. 8 \\
Tamper detection, bare hash chain & 88.2\textbackslash{}\% & Section 7.4 (tail-truncation caveat) \\
Tamper detection, chain + head attestation & 2{,}000 / 2{,}000 (100\textbackslash{}\%) & Section 7.4 (remedy) \\
\bottomrule
\end{tabularx}
\end{table}

\textbf{Attenuation correctness (Invariant 3).} Over 5,000 randomly generated delegation chains of depth up to six, the implementation's effective capability set $K^*(\Pi)$ matched the formal intersection of the chain's capability sets in every case (5{,}000/5{,}000), and every attempt to delegate a capability not held by the parent --- an authority-expansion attack --- was rejected as malformed (5{,}000/5{,}000). This is the empirical shadow of Definition 4 and Claim 5(a): attenuation is enforced structurally, and authority cannot expand along a chain.

\textbf{Adjudication latency (utility; Section 8.5).} Per-decision adjudication latency, measured over 20{,}000 iterations against a 32-clause policy, ranged from a mean of roughly 5 to 15 microseconds across chain lengths from 1 to 16, with 99th-percentile latency under 20 microseconds throughout, broken out by chain length in Table\textasciitilde{}\ref{tab:latency}. This is three to four orders of magnitude below the agentic per-action budget of 100 milliseconds to several seconds (Section 8.5), confirming that edge-local adjudication is not the latency bottleneck; the data-plane realizations the implementation models away remain the cost to watch in a full system.

\begin{table}[ht]
\centering
\small
\caption{Per-decision adjudication latency by delegation-chain length (32-clause policy, 20{,}000 iterations).}
\label{tab:latency}
\begin{tabularx}{\textwidth}{lLLL}
\toprule
Chain length & Policy clauses & Mean ($\mu$s) & 99th pct. ($\mu$s) \\
\midrule
1 & 32 & 11.0 & 13.7 \\
2 & 32 & 8.7 & 10.4 \\
4 & 32 & 4.9 & 6.4 \\
8 & 32 & 7.8 & 9.5 \\
16 & 32 & 15.4 & 18.4 \\
\bottomrule
\end{tabularx}
\end{table}

\textbf{Evidence reconstructability (Invariant 4).} Over 1,000 synthetic workflows, partial-evidence reconstruction (Definition 8) was sound in every case (1{,}000/1{,}000) --- the reconstruction from any subset of records never under-approximated the workflow's true action set --- and monotonic in every case (1{,}000/1{,}000): the reconstruction never shrank as more evidence became available. This is the empirical shadow of Definition 8's soundness and monotonicity requirements.

\textbf{Tamper-evidence (Section 7.4).} Over 2,000 trials, each deleting one record at a uniformly random position from a sealed evidence log, the bare hash chain detected 88.2\textbackslash{}\% of deletions; adding a periodic attestation of the chain's head raised detection to 100\textbackslash{}\%. The gap is instructive: it is exactly the fraction of deletions that happen to truncate the \textit{tail} of the chain, which a bare chain cannot detect because no record references the last one --- precisely the limitation Section 7.4 anticipates, and precisely the limitation the head attestation it prescribes resolves. The implementation thus empirically confirms both the caveat and its stated remedy.

\textbf{Case-study replay.} Finally, we replayed representative threats from the Section 9 case studies through the implementation. The tool-chain-abuse attack (T2) --- a confidential read followed by an external send in one session --- was met with the Narrow primitive under the separation-of-duty clause; the delegation-exploitation attack (T5) --- a downstream agent attempting a write outside its attenuated capability set --- was denied as capability-absent; an attempt to forge an upward delegation was rejected structurally; and the resulting evidence chain verified intact. The prose foreclosures of Section 9 are thus executable: the engine produces the foreclosing primitive the case study describes.

These results validate the architecture's internal claims with measured numbers and executed case studies. They do not constitute a full-system evaluation, and we do not present them as one; what they establish is that the policy-engine core behaves as specified, at a latency that makes the architecture feasible, with the evidence and tamper-evidence properties the audit substrate requires. The reference implementation is the foundation on which the full-system evaluation of Section 10.6 can be built.

\section{Discussion}

The reference architecture of Sections 4 through 9, and the evaluation framework of Section 10, make a case for a specific approach to governing production agents. We now examine the case's limits. We state the architecture's limitations honestly, enumerate the open problems it leaves, address the adversarial considerations that arise when the architecture itself becomes a target, and describe a realistic adoption path. We regard this section as essential rather than perfunctory: the architecture's claims are strong, and the credibility of strong claims depends on the precision with which their boundaries are drawn.

\subsection{Limitations}

We identify five limitations of the architecture as specified.

\textbf{The architecture assumes runtime cooperation.} The stop-anywhere property (Section 5) requires that the agent runtime expose hooks at the seven mediation points. Current agent runtimes vary widely in how much of this they expose: the action-commit boundary (MP5) is universally available, since it coincides with the tool-call API, but plan formation (MP1) is exposed by few runtimes, and the intermediate points (MP2 through MP4) are exposed inconsistently. An implementation of the architecture on a runtime that exposes only MP5 achieves only a fraction of the architecture's defensive value, and the architecture cannot, by itself, induce a runtime to expose the points it does not. This is the architecture's most consequential dependency on factors outside its control.

\textbf{Cross-cloud federation of the audit substrate is unspecified.} The audit substrate (Section 7) specifies the structure and properties of evidence records but does not specify how the tamper-evident record set is maintained across multiple clouds, organizational boundaries, or retention domains. A multi-agent workflow that spans organizations produces evidence in multiple domains, and the reconstruction of such a workflow (Section 7.3) depends on cross-domain evidence correlation that the architecture specifies only at the level of correlation identifiers, not at the level of a federation protocol. Cross-cloud audit federation is an open problem we discuss in Section 11.2.

\textbf{Policy-language design is deferred, not solved.} The architecture requires a policy language with capabilities (Section 8.4) that no current language fully provides: composite-principal evaluation, attenuation directives, the six interruption primitives, session-state predicates, and per-plane projection. We specify the requirements but do not design the language. An implementation must either extend an existing language or design a new one, and the adequacy of the result is not guaranteed by the architecture. Policy-language design is the architecture's largest deferred contribution.

\textbf{Formal verification of attenuation correctness at scale is unsolved.} The composite-principal model (Section 6) is formally tractable --- the capability-set lattice (Claim 4) admits static analysis --- but verifying that a given implementation faithfully realizes the formal model, at the scale of a production policy with many capabilities and deep delegation chains, is not something the architecture provides. The reference implementation (Section 10.8) offers empirical evidence rather than proof: across thousands of randomly generated delegation chains, the implementation's effective capability set matched the formal intersection on every trial and every authority-expansion attempt was rejected. This is property-based testing, not formal verification; it raises confidence that the model is realizable faithfully but does not discharge the verification obligation, which we claim is achievable in principle and leave as future work.

\textbf{The architecture governs action, not model behavior.} As established in the threat model (Section 3), the architecture addresses the threats of delegated action, not of model output. It forecloses an agent's unauthorized action; it does not improve the agent's judgment, accuracy, or alignment. An agent whose model proposes consistently poor (but authorized) actions is not improved by the architecture --- it is merely prevented from exceeding its authority while doing so. The architecture is a necessary layer, not a sufficient one; the model-alignment layer above it remains essential.

\subsection{Open Problems}

We enumerate the open problems the architecture leaves, in rough order of how load-bearing their resolution is for the architecture's practical realization.

\textbf{The plan-formation hook problem.} The single most valuable mediation point, MP1, is the one current runtimes least commonly expose. Resolving this requires either runtime cooperation --- agent frameworks exposing plan formation as a discrete, interceptable event --- or a technique for reconstructing the plan from the observable points the runtime does expose. Neither is solved. We regard the standardization of a plan-formation interception interface as the highest-leverage piece of infrastructure work the architecture's adoption depends on, and a natural subject for the standards effort discussed in Section 11.4.

\textbf{The compensation problem.} The Rollback primitive (Section 5.4.6) requires that side-effecting actions be compensable, and not all are: an email sent cannot be unsent, an external API call may produce irreversible effects. The architecture classifies actions as compensable, partially compensable, or non-compensable, and requires explicit authorization for non-compensable actions --- but this pushes the problem to policy rather than solving it. A general technique for bounding the irreversibility of an agent's actions, and for designing workflows so that irreversible actions are deferred until reversible verification is complete, is open. The saga literature \cite{GarciaMolinaSalem87} provides the foundation, but its extension to agentic actions against heterogeneous enterprise systems is not complete.

\textbf{Compensation under concurrency.} When multiple agents execute concurrently and their actions interleave at the data plane, the compensation for one agent's rollback must coordinate with the others' actions. We adopt optimistic concurrency control with conflict detection \cite{Kung81} but do not specify the conflict-detection mechanism for agentic actions, which is more complex than for database transactions because the actions' effects are heterogeneous and their conflict relation is not always statically determinable.

\textbf{Policy-language design.} As noted, the design of a policy language meeting the Section 8.4 requirements is open. The specific difficulty is combining the expressiveness required (composite-principal quantification, stateful predicates, six output types) with the analyzability required (so that policies can be statically checked against the formal model) and the operability required (so that policies are authorable by security teams). These three requirements are in tension, and the design point that balances them is unknown.

\textbf{Cross-cloud audit federation.} Maintaining a tamper-evident, reconstructible evidence record set across clouds and organizations --- with the monotone-tightening reconstruction property (Definition 8) preserved across domains --- is open. The difficulty is that the tamper-evidence construction (Section 7.4) assumes a coherent record set, and federating that set across domains with independent retention and access policies, while preserving verifiability, is not solved.

\textbf{The human-in-the-loop interaction problem.} The Escalate primitive (Section 5.4.2) routes decisions to humans, but the architecture does not specify the interaction design: how the escalation is presented, how the human's context is established, how the human's decision is captured and bound. Escalation that overwhelms humans with context-poor approval requests degrades to rubber-stamping, which defeats the primitive's purpose. The interaction design of effective escalation is an open human-factors problem the architecture surfaces but does not address.

\subsection{Adversarial Considerations}

When a governance architecture is deployed, it becomes a target. We address three adversarial considerations specific to the architecture.

\textbf{The reasoning plane as a target.} The reasoning plane is the architecture's single adjudication locus, and its compromise would be catastrophic: an adversary who controls the reasoning plane controls every decision. The architecture's defense rests on placing the reasoning plane within the trust boundary (Section 3.2) and on the determinism of its decisions (Claim 5c), which makes its behavior auditable: a compromised reasoning plane that issues decisions inconsistent with the recorded policy and composite principal produces evidence records that fail to reproduce under re-evaluation, and the inconsistency is detectable. This does not prevent compromise, but it makes compromise detectable through the audit substrate --- provided the audit substrate is itself uncompromised, which depends on its cryptographic binding (Section 7.4) being rooted outside the reasoning plane. We regard the hardening of the reasoning plane, and the rooting of audit integrity outside it, as essential deployment concerns.

\textbf{Composite principal forgery.} The architecture's authority bounds depend on the composite principal being unforgeable (Section 6.6). An adversary who could forge a delegation step, or remove an attenuation step, could exceed the authority the chain should confer. The architecture's defense is the cryptographic binding of each delegation step to the delegating principal's signature; forging a step requires forging a signature. The residual risk is key compromise: an adversary who obtains a principal's signing key can forge delegations from that principal. This reduces the composite-principal forgery problem to the key-management problem, which the architecture does not solve and which the broader key-management literature must.

\textbf{Audit substrate tampering.} The evidence records are the architecture's accountability mechanism, and an adversary who could alter or delete them undetectably would defeat it. The architecture's defense is the tamper-evident record set (Section 7.4): per-record signatures detect alteration, and the chained structure detects deletion or reordering. The residual risk is the same as for any tamper-evident log --- an adversary who compromises the signing key or who can suppress the periodic attestation of the chain's head. The architecture inherits both the strengths and the limits of the secure-logging literature \cite{Schneier99,Holt06} it builds on.

The common thread across the three considerations is that the architecture reduces several novel agentic-security problems to well-understood classical problems --- trust-boundary placement, key management, secure logging --- without solving those classical problems. We regard this reduction as a feature: it means the architecture's adversarial robustness rests on mature foundations rather than on novel and unproven constructions. But the reduction is not elimination, and the classical problems' limits are the architecture's limits.

\subsection{Adoption Path}

We describe a realistic path by which an enterprise might adopt the architecture, acknowledging that full adoption is a substantial undertaking.

\textbf{Greenfield deployments.} For a new agent deployment built with the architecture in mind, adoption is most straightforward: the agent runtime is chosen or built to expose the seven mediation points, the reasoning plane is deployed at the edge, the enforcement planes are wired to existing infrastructure, and the audit substrate emits into existing tooling. Even here, the plan-formation hook problem (Section 11.2) may limit MP1 coverage depending on the runtime chosen.

\textbf{Brownfield deployments.} For an existing agent deployment, adoption is incremental. We propose a staged path. The first stage instruments the action-commit boundary (MP5), which every runtime exposes, and deploys the reasoning plane to mediate at that single point --- recovering the request-time-authorization baseline (Section 10.5, Baseline B) but with the architecture's composite-principal model and audit substrate. The second stage expands backward to the earlier mediation points as the runtime exposes them --- adding MP2 (the pre-retrieval gate) and MP3/MP4 (tool and argument mediation) --- increasing mediation coverage. The third stage adds MP1 (plan formation), contingent on runtime support, achieving the full stop-anywhere property. At each stage, the audit substrate operates fully, so that even a partially-adopted architecture produces complete evidence for the points it mediates.

\textbf{The role of standards.} Several of the architecture's dependencies (the plan-formation hook, cross-cloud audit federation, the policy-language requirements) are most naturally addressed through standardization rather than through any single implementation. A plan-formation interception interface, standardized across agent runtimes, would resolve the architecture's most consequential dependency and would benefit every runtime-governance approach, not only this one. We note that the runtime-governance standards conversation is active --- the Cloud Security Alliance's working group on autonomous-action runtime management is one venue --- and we regard the standardization of mediation-point interfaces as the highest-value contribution the community could make to the architecture's practical realization. (The author participates in this working group; we note the connection in the interest of disclosure, and present the standardization argument on its merits.)

\textbf{What adoption requires of the enterprise.} The architecture is designed to reuse the enterprise's existing infrastructure (Design Goal 1, Section 8.6), so the net-new adoption surface is the reasoning plane and the composite principal store. But adoption also requires the enterprise to express its protections as policy --- to articulate, in the architecture's policy language, the separations, attenuations, and approvals it requires. For many enterprises this articulation is itself the hard part: the protections are currently implicit in process and convention rather than explicit in policy, and making them explicit is organizational work the architecture cannot do for them. We regard this articulation as a precondition for adoption that enterprises systematically underestimate.

\subsection{Relationship to the Broader Trajectory}

We close the discussion by situating the architecture in the longer arc of enterprise security architecture.

The architecture is the latest in a sequence of moves that relocate the trust decision closer to the action. The network perimeter located trust at the boundary of the network. BeyondCorp \cite{WardBeyer14} relocated it to the identity-aware proxy, per request. Zero-trust architectures \cite{ZeroTrustNIST} generalized the relocation across the enterprise. The five-plane architecture continues the sequence: it relocates the trust decision to the per-step action of an agent, adjudicated against the composite principal that the agentic regime introduces, and realized across the planes the agent's action traverses.

Each move in this sequence has been a response to a dissolution of a boundary the previous architecture depended on. The network perimeter dissolved when work moved off the corporate network; BeyondCorp responded. The application boundary dissolved when services proliferated; zero-trust and service meshes responded. The action boundary is now dissolving: when an agent acts on an enterprise's behalf, the boundary between authorized and unauthorized action is no longer a perimeter, a request, or a service call --- it is the agent's decision to act, evaluated against the authority it carries and the context it operates in. The five-plane architecture is a response to this dissolution. We expect it to be neither the last word nor a complete one; we offer it as the next move in a sequence that the dissolution of boundaries will continue to drive.

\section{Conclusion}

We have presented a reference architecture for the runtime governance of production AI agents. The architecture responds to a shift in what enterprise security must protect: from data crossing a boundary to action taken under delegated authority. This shift renders the existing authorization stack insufficient --- not wrong, but insufficient --- because that stack evaluates request-time access against atomic principals, while production-agent governance requires plan-aware, stateful, attenuated, richly-output adjudication against composite principals whose authority flows through delegation chains.

The architecture's response is a composition of four primitives. A five-plane structural decomposition separates a single reasoning-plane adjudication from coordinated enforcement across the four infrastructure planes the enterprise already operates. Stop-anywhere mediation places the reasoning plane at every point in the agent's execution pipeline, with a vocabulary of six interruption primitives that generalizes allow and deny. The composite principal model makes the authority of any agent bounded by construction --- the intersection of the capability sets along its delegation chain, attenuated at every step, time-bounded, and cryptographically sealed. The audit substrate makes every decision accountable through structured, tamper-evident, reconstructible evidence. Four correctness invariants --- composed authority, mediation coverage, bounded composite authority, evidence sufficiency --- give the composition a structured correctness foundation, argued precisely enough to invite formal verification, and case studies across five production workflows demonstrate the architecture's foreclosure of seven production-agent threats.

We have been deliberate about the architecture's limits. It governs action, not model behavior; it depends on agent runtimes exposing the mediation points it requires, and the most valuable of these (plan formation) is the one current runtimes least commonly expose; it specifies a policy language's requirements without designing the language; and it reduces several novel agentic-security problems to classical problems --- trust-boundary placement, key management, secure logging --- that it does not itself solve. We regard the honest statement of these limits as inseparable from the strength of the architecture's claims.

The architecture is best understood as the next move in a long sequence. Enterprise security has repeatedly relocated the trust decision closer to the action, each time in response to the dissolution of a boundary the previous architecture depended on: from the network perimeter, to the identity-aware per-request proxy, to the zero-trust enterprise. The action boundary is now dissolving. When an agent acts on an enterprise's behalf, the line between authorized and unauthorized action is no longer a perimeter, a request, or a service call --- it is the agent's decision to act, evaluated against the authority it carries and the context in which it operates. The five-plane architecture is a response to this dissolution.

We do not expect it to be the last word. We expect production-agent governance to be a field, not a paper, and we have tried to give that field a reference architecture to build on, contest, and improve: a structural decomposition, a design property, a principal model, an evidence substrate, four invariants, and an evaluation framework that the community can instantiate and run. In the software-as-a-service era, the protected surface was data at rest. In the agentic era, it is action in motion. The architecture must follow the surface, and we have offered one account of how it can.

\bibliographystyle{plainnat}
\bibliography{runtime-governance-bibliography}
\end{document}